\documentclass[]{dataflow}

\usepackage[toc,page,header]{appendix}

\usepackage{minitoc}
\usepackage{multirow}     % \multirow
\usepackage{multicol}
\usepackage{booktabs}     % \toprule \midrule \bottomrule
\usepackage{colortbl}
\usepackage[table]{xcolor}
\usepackage{makecell}
\usepackage{graphicx}     % \resizebox
\usepackage{enumitem}
\usepackage{amsmath,amssymb}
\usepackage{url}
\usepackage{fontawesome5}

\definecolor{model_type}{RGB}{224,225,221}
\definecolor{column_green}{RGB}{223,237,239}
\definecolor{column_yellow}{RGB}{255,239,189}

\usepackage[most]{tcolorbox}
\definecolor{prompt_yellow}{RGB}{249,240,200}
\definecolor{prompt_green}{RGB}{219,231,231}
\definecolor{prompt_blue}{RGB}{186,216,240}
\definecolor{prompt_purple}{RGB}{219,205,240}
\definecolor{frame1}{RGB}{111,135,141}
\definecolor{frame2}{RGB}{69,82,129}
\definecolor{frame3}{RGB}{121,117,99}
\definecolor{frame4}{RGB}{115,93,120}

\title{K12-KGraph: A Curriculum-Aligned Knowledge Graph for Benchmarking and Training Educational LLMs}

\author[*,1,4]{Hao Liang}
\author[*,1]{Qihan Lin}
\author[2,4]{Mingrui Chen}
\author[1]{Hengyi Feng}
\author[1]{Zhaoyang Han} 
\author[3]{Xiaochen Ma} 
\author[1]{Zhen Hao Wong} 
\author[1]{Meiyi Qiang} 
\author[2]{Linzhuang Sun} 
\author[\ddagger,1,4]{Wentao Zhang}

\affiliation[]{$^{1}$Peking University, $^{2}$University of Chinese Academy of Sciences $^{3}$HKUST $^{4}$Zhongguancun Academy}

\contribution[*]{Equal Contribution}
\contribution[\ddagger]{Corresponding author}

\abstract{
Large language models (LLMs) are increasingly deployed in K--12 education, yet existing benchmarks such as C-Eval, CMMLU, GaokaoBench, and EduEval measure only whether a model can \emph{answer} an exam question, i.e., factual recall. Effective educational AI further requires \textbf{curriculum cognition}: the structured understanding of how knowledge is organized and visually presented, including prerequisite chains, concept taxonomies, experiment--concept links, and pedagogical sequencing. Curriculum cognition is neither probed by current benchmarks nor explicitly taught by current instruction-tuning data. To close this gap, we introduce \textbf{K12-KGraph}, a curriculum-aligned knowledge graph extracted from the official People's Education Press textbooks, covering mathematics, physics, chemistry, and biology across primary, middle, and high school, with nine node types (\texttt{Book},  \texttt{Chapter}, \texttt{Section},\texttt{Concept}, \texttt{Skill}, \texttt{Experiment}, \texttt{Exercise},  \texttt{Figure}, \texttt{VisualElement}) and fourteen relation types spanning both curriculum structure and visual grounding. Building on this single graph, we derive two complementary resources: \textbf{K12-Bench}, a 23{,}640-question multi-select benchmark across five graph-derived task families (\textsc{Ground}, \textsc{Prereq}, \textsc{Neighbor}, \textsc{Evidence}, and \textsc{Locate}) that jointly probe curriculum cognition; and \textbf{K12-Train}, a KG-guided supervised fine-tuning corpus of 7{,}335 samples (K12-Train-Full), including 2{,}267 text-only QA pairs (K12-Train-Text) and 5{,}068 multimodal VQA pairs (K12-Train-MM). 
Experiments expose a clear gap and a clear remedy: on K12-Bench, even a strong proprietary model (Gemini-3-Flash) reaches only $57\%$ exact match and a strong open-source model (Gemma-4-31B-IT) only $46\%$, with \textsc{Prereq} and \textsc{Neighbor} being the hardest; yet our training experiments show that domain-specific supervision from K12-Train can effectively address this limitation.
For LLMs, under a strictly matched 2{,}300-sample SFT budget, K12-Train-Text consistently outperforms equally sized subsets of eight mainstream instruction-tuning corpora (OpenHermes, Infinity, UltraChat, WizardLM, DataFlow, LMSYS, SmolTalk, Tulu-3) on both GaokaoBench and EduEval, demonstrating the sample efficiency of structurally grounded curriculum data. 
For VLMs, K12-Train-Full achieves the best overall performance on Gaokao-MM, MDK12-medium, and K12Vista among all compared training configurations despite using fewer samples than the full DataFlow and WizardLM baselines, and consistently outperforms both K12-Train-Text and K12-Train-MM, showing the complementarity of textual and visual supervision. 
}
\date{\today}

\def\githubicon{\raisebox{-1.5pt}{\includegraphics[height=1.05em]{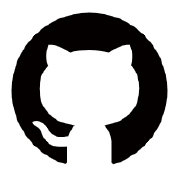}}}
\def\huggingfaceicon{\raisebox{-1.5pt}{\includegraphics[height=1.05em]{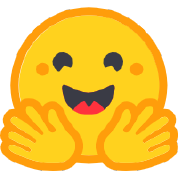}}}

\newcommand{\sourcelink}{https://github.com/haolpku/K12-Dataset}
\newcommand{\datalink}{https://huggingface.co/datasets/lhpku20010120/K12-KGraph}

\checkdata[ \githubicon \hspace{0.3em} Source Code ]{ \url{\sourcelink} }
\checkdata[ \huggingfaceicon \hspace{0.3em} Dataset ]{ \url{\datalink} }
\checkdata[ \faFile \hspace{0.57em} Dataset Website ]{ \url{https://haolpku.github.io/K12-KGraph-page/} }

\begin{document}
\maketitle

% footnote
\renewcommand{\thefootnote}{\fnsymbol{footnote}}
\setcounter{footnote}{0}

\renewcommand{\thefootnote}{\arabic{footnote}}
\pagestyle{fancy}
\fancyhf{}
\fancyhead[L]{K12-KGraph}
\fancyhead[R]{\thepage}

% Content
\newpage
\tableofcontents
\newpage

\section{Introduction}
\label{sec:intro}

Large language models (LLMs) have become strikingly proficient at answering K--12 exam questions. On benchmarks such as C-Eval~\citep{huang2024ceval}, CMMLU~\citep{li2023cmmlu}, GaokaoBench~\citep{zhang2023gaokao}, and EduEval~\citep{ma2025edueval}, frontier models now rival, and sometimes surpass, top human students, fueling rapid interest in LLM-powered tutoring and exam preparation~\citep{kasneci2023chatgpt}. Taken at face value, this progress suggests that educational AI is close to being solved. Yet anyone who has tried to build a real tutoring product knows that \emph{answering} a question is only a small fraction of what a good teacher actually does, and it is precisely the larger fraction that today's benchmarks leave untested.

The untested part is what we call \textbf{curriculum cognition}: the structured understanding of \emph{why} a topic must be learned before another, \emph{how} a laboratory experiment connects to a theoretical concept, and \emph{where} in the textbook each idea actually lives. A competent 7th-grade mathematics teacher does not merely know that ``linear equations'' is a topic; she knows it requires arithmetic operations as a prerequisite, that it sits as a sibling of ``inequalities'' under ``algebraic expressions'', and that it first appears in Chapter~3 of the People's Education Press textbook. Curriculum knowledge is also communicated through textbook figures: diagrams, experimental setups, geometric figures, and exercise illustrations can explain a concept, provide visual evidence for a relation, or supply information required to solve a problem. None of today's K--12 benchmarks probe this kind of structural and visual understanding, and consequently none of today's training data explicitly teaches it either. If we care about building AI that supports students, rather than merely testing them, this gap is the bottleneck.

To close it, we build everything from a single resource: a curriculum-aligned \emph{knowledge graph} extracted directly from the official Chinese K--12 textbooks (People's Education Press). The resulting graph, \textbf{K12-KGraph}, is a heterogeneous property graph covering mathematics, physics, chemistry, and biology across primary, middle, and high school. 
It comprises two components: a textual component that captures curriculum structure and a multimodal component that captures visual grounding.
The textual component contains seven node types (\texttt{Concept}, \texttt{Skill}, \texttt{Experiment}, \texttt{Exercise}, \texttt{Section}, \texttt{Chapter}, and \texttt{Book}), and edge types encoding taxonomy (\texttt{is\_a}), prerequisite (\texttt{prerequisites\_for}), associations (\texttt{relates\_to}), verification (\texttt{verifies}), assessment (\texttt{tests\_concept}, \texttt{tests\_skill}), location (\texttt{appears\_in}), and order (\texttt{leads\_to}).
The multimodal component contains two node types (\texttt{Figure} and \texttt{VisualElement}) and edge types encoding composition (\texttt{contains\_visual\_element}), visual semantics (\texttt{refers\_to}, \texttt{illustrates}), location (\texttt{appears\_in}), exercise dependence on figures (\texttt{requires\_figure}), and visual evidence (\texttt{supports\_edge}). 
Because the graph faithfully mirrors how the curriculum is organized, we can turn it into evaluation questions by traversing neighborhoods, and into training data by rendering node properties and edge semantics into QA pairs. A single graph thus yields both a \emph{benchmark} that measures curriculum cognition and a \emph{training set} that explicitly teaches it.

\paragraph{Contributions.} Our work makes three contributions:
\begin{enumerate}[nosep,leftmargin=*]
    \item \textbf{K12-KGraph}, a large-scale, multi-subject, official-textbook-grounded curriculum-aligned knowledge graph for Chinese K--12, together with a reproducible LLM-based extraction and hierarchical-merge pipeline with DAG validation on taxonomic and prerequisite relations.
    \item \textbf{K12-Bench}, a 23{,}640-question benchmark of graph-derived multi-select items grouped into five task families: \textsc{Ground} (Knowledge Grounding), \textsc{Prereq} (Prerequisite Reasoning), \textsc{Neighbor} (Neighbor Recommendation), \textsc{Evidence} (Experiment Evidence Chain), and \textsc{Locate} (Cross-Chapter Indexing). These families together probe structural curriculum understanding. Evaluating ten open-source and proprietary LLMs, we find that even Gemini-3-Flash reaches only 57\% exact match, and a strong open-source model (Gemma-4-31B-IT~\citep{team2024gemma}) reaches only 46\%.
    \item \textbf{K12-Train}, a KG-guided QA synthesis pipeline yielding 7{,}335 high-quality educational SFT samples, including a K12-Train-Text containing 2{,}267 text-only samples, and a K12-Train-MM containing 5{,}068 multimodal samples. For LLMs, under a strictly matched 2{,}300-sample budget, SFT on K12-Train-Text consistently outperforms eight mainstream instruction-tuning corpora (OpenHermes~\citep{OpenHermes25}, Infinity~\citep{li2025infinity}, UltraChat~\citep{ding2023enhancing}, WizardLM~\citep{luo2023wizardcoder}, DataFlow~\citep{liang2025dataflow}, LMSYS~\citep{zheng2023lmsys}, SmolTalk~\citep{allal2025smollm2}, Tulu-3~\citep{lambert2024tulu}) across two base models on GaokaoBench and EduEval (e.g., $+24.1$/$+32.4$ over the strongest SFT baseline and $+114.6$/$+221.0$ over official instruction-tuned variants on GaokaoBench). 
    % On GaokaoBench, K12-Train (Text Only) achieves the best overall performance across both backbones, improving over the strongest SFT baseline by $+24.1$ (Qwen3-4B-Base) and $+32.4$ (Llama3.1-8B-Base), and over their corresponding official instruction-tuned variants by $+114.6$ and $+221.0$, respectively. On EduEval, it also attains the highest average score among all SFT configurations, indicating that KG-grounded in-domain supervision is highly effective for educational SFT. 
    For VLMs, SFT on K12-Train-Full achieves the best overall performance on Gaokao-MM, MDK12-Bench, and K12Vista, surpassing the full DataFlow and WizardLM baselines with a much smaller sample size (7{,}335 vs.\ 10{,}000/142{,}759), as well as the performance of K12-Train-Text and K12-Train-MM.
    These results highlight both the value of in-domain educational data and the complementarity of textual and visual supervision.
\end{enumerate}

\section{Related Work}
\label{sec:related}

\paragraph{K--12 and education benchmarks.}
Several benchmarks evaluate LLMs on Chinese K--12 subjects. C-Eval~\citep{huang2024ceval} and CMMLU~\citep{li2023cmmlu} are broad multi-discipline suites that include K--12 categories but focus on multiple-choice factual questions. GaokaoBench~\citep{zhang2023gaokao} targets the Chinese college entrance exam (Gaokao) with both objective and subjective questions. EduEval~\citep{ma2025edueval} covers six educational capability dimensions (application, creativity, ethics, memory, reasoning, understanding). E-Eval~\citep{yu2024eeval} and K--12 EduBench~\citep{ye2026k} further expand coverage. Multimodal benchmarks such as Gaokao-MM~\citep{zong2024gaokao}, MDK12-Bench~\citep{zhou2026mdk12}, and K12Vista~\citep{li2025k12vista} evaluate visual reasoning over educational questions. CK12~\citep{you2024ck12} incorporates a knowledge graph but focuses on holistic cognition rather than structural curriculum understanding. All of these benchmarks test whether a model can answer domain questions; none systematically evaluate whether models understand the \emph{structure} of the curriculum, i.e., prerequisite dependencies, concept taxonomies, or pedagogical sequencing.

\begin{figure*}[t]
  \centering
  \includegraphics[width=1\linewidth]{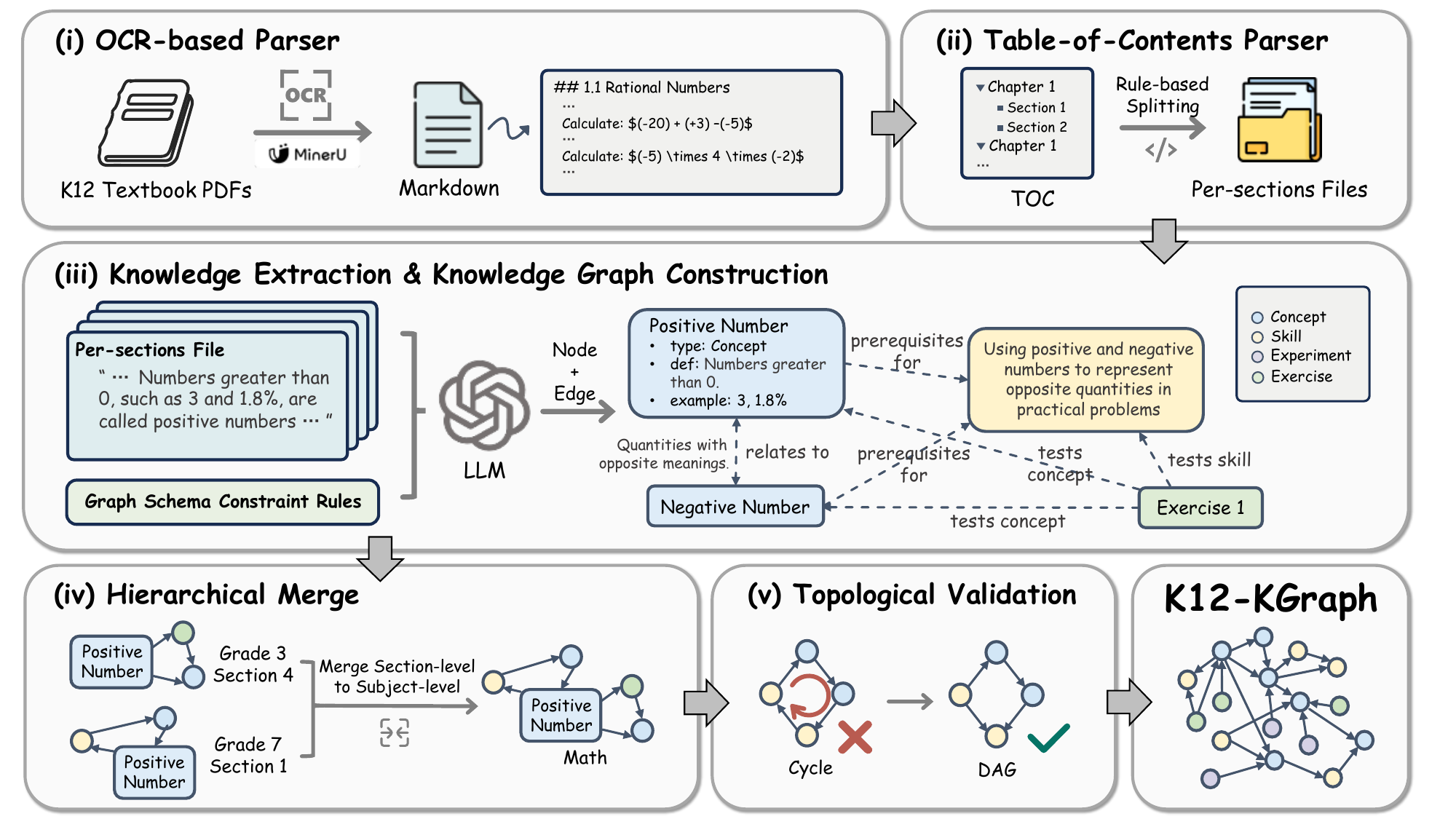}
  \caption{\textbf{Overview of the K12-KGraph construction pipeline.} The process consists of five stages: OCR-based parsing of textbooks, hierarchical segmentation into sections, LLM-based schema-guided extraction of nodes and edges, hierarchical graph merging across sections and books, and structural validation with lightweight human verification. The design combines automated extraction with rule-based processing and human-in-the-loop checks to ensure both scalability and correctness.}
  \label{fig:KGraph}
\end{figure*}

\paragraph{Educational knowledge graphs.}
Knowledge graphs have been applied in education for knowledge tracing~\citep{liu2019ekt} and prerequisite discovery~\citep{chen2018prerequisite}. However, existing educational KGs typically focus on a single subject or English-language courses, and are rarely aligned to an official K--12 curriculum, and generally represent textbook knowledge as textual entities and relations without explicitly grounding textbook figures. Recent work has explored using LLMs for KG construction~\citep{wei2023zeroshot, zhu2024llms4ol}, but no prior work has built a large-scale, multi-subject, curriculum-aligned KG from Chinese K--12 textbooks and used it to both benchmark and train LLMs.

\paragraph{Instruction tuning datasets.}
Supervised fine-tuning (SFT) with high-quality instruction data is critical for aligning LLMs~\citep{ouyang2022training}. General-purpose datasets such as OpenHermes~\citep{OpenHermes25}, UltraChat~\citep{ding2023enhancing}, WizardLM~\citep{luo2023wizardcoder}, Tulu-3~\citep{lambert2024tulu}, and SmolTalk~\citep{allal2025smollm2} cover diverse tasks but lack domain-specific educational knowledge. Self-Instruct~\citep{wang2023selfinstruct} and its variants generate synthetic data from LLMs, but do not leverage structured knowledge sources. Our KG-guided synthesis approach bridges this gap by grounding training data in curriculum structure.

\begin{table*}[t]
\centering
\caption{\textbf{Statistics of the textual component of K12-KGraph by subject.}
We report the selected curriculum-content node types and relation types after hierarchical merging and deduplication. The total number of \texttt{leads\_to} edges is computed globally and includes cross-subject links, and therefore does not equal the sum of the subject-specific rows.}
\label{tab:kg_stats}
\small
\setlength{\tabcolsep}{3pt}
\resizebox{\linewidth}{!}{
\begin{tabular}{@{}lrrrrr|rrrrrrr@{}}
\toprule
\multirow{2}{*}{\textbf{Subject}} & \multicolumn{5}{c|}{\textbf{Nodes}} & \multicolumn{7}{c}{\textbf{Edges}} \\
& \texttt{Book} & \texttt{Cpt} & \texttt{Skl} & \texttt{Exp} & \texttt{Exe} & \texttt{is\_a} & \texttt{prereq} & \texttt{rel\_to} & \texttt{verif} & \texttt{tes\_cpt} & \texttt{tes\_skl} & \texttt{lea\_to} \\
\midrule
Mathematics & 23 & 1{,}475 & 428 & 0   & 471 & 288 & 855 & 405 & 0   & 464 & 229 & 44 \\
Physics     & 9  & 1{,}154 & 197 & 220 & 186 & 247 & 648 & 300 & 251 & 248 & 104 & 19 \\
Chemistry   & 7  & 2{,}302 & 451 & 309 & 270 & 856 & 1{,}344 & 1{,}083 & 468 & 571 & 251 & 21 \\
Biology     & 9  & 1{,}648 & 288 & 123 & 244 & 505 & 858 & 386 & 170 & 391 & 116 & 11 \\
\midrule
\textbf{Total} & 48 & 6{,}579 & 1{,}364 & 652 & 1{,}171 & 1{,}896 & 3{,}705 & 2{,}174 & 889 & 1{,}674 & 700 & 282 \\
\bottomrule
\end{tabular}
}
\vspace{1mm}
\end{table*}

\begin{table}[t]
\centering
\caption{\textbf{Statistics of the multimodal component of K12-KGraph by subject.} Counts are reported after hierarchical merging and deduplication.}
\label{tab:kg_stats_mm}
\small
\setlength{\tabcolsep}{3pt}
\begin{tabular}{@{}lrr|rrrrr@{}}
\toprule
\multirow{2}{*}{\textbf{Subject}} & \multicolumn{2}{c|}{\textbf{Nodes}} & \multicolumn{5}{c}{\textbf{Edges}} \\
& \texttt{Fig} & \texttt{Vis\_Ele} & \texttt{con\_ve} & \texttt{ref\_to} & \texttt{illus} & \texttt{req\_fig} & \texttt{sup\_edge} \\
\midrule
Mathematics & 3,831 & 5,208 & 4,490 & 5,256 & 4,006 & 59 & 529 \\
Physics     & 1,344 & 1,906 & 1,738 & 1,927 & 1,407 & 0  & 365 \\
Chemistry   & 1,096 & 1,583 & 1,485 & 1,642 & 1,154 & 43 & 568 \\
Biology     & 1,117 & 1,506 & 1,364 & 1,527 & 1,152 & 0  & 369 \\
\midrule
\textbf{Total} & 7,388 & 10,203 & 9,077 & 10,352 & 7,719 & 102 & 1,831 \\
\bottomrule
\end{tabular}
\vspace{1mm}
\end{table}

\section{K12-KGraph}
\label{sec:kgraph}

\subsection{Schema Design}
\label{sec:schema}
We design a heterogeneous property graph schema tailored to K--12 education, which includes both textual and multimodal components. The schema defines nine node types and fourteen directed edge types, summarized in Table~\ref{tab:schema} (Appendix~\ref{app:field_schema}).

Each node carries typed properties. For example, a \texttt{Concept} node includes \texttt{name}, \texttt{definition} (preferring textbook wording), \texttt{importance} (understand/master/important), and optional fields such as \texttt{formula}, \texttt{aliases}, and \texttt{examples}. An \texttt{Experiment} node includes \texttt{instruments}, \texttt{is\_student} (whether it is a student-performed experiment), \texttt{process}, \texttt{phenomena}, and \texttt{conclusion}.

\subsection{Construction Pipeline}
\label{sec:pipeline}

Construction proceeds in five automatic stages, as illustrated in Figure~\ref{fig:KGraph}. (i)~An OCR-based parser (MinerU~\citep{niu2025mineru2}) converts textbook PDFs into structured Markdown while preserving heading hierarchy, mathematical formulas, raw text, and image assets. (ii)~A table-of-contents parser produces a \texttt{sections\_index.json} manifest and splits the Markdown into per-section files, and associates each image with its section text. (iii)~For each section, we prompt GPT-5.2 with a schema-aware instruction (Appendix~\ref{app:prompts}) to emit nodes and edges as structured JSON, together with evidence citations (original sentences from textbooks) or confidence scores (reflecting model's confidence in the relation) for every edge. (iv)~Per-section graphs are merged bottom-up: a book-level pass assigns globally unique IDs, deduplicates same-name concepts and skills, and remaps edges; a subject-level pass then reconciles entities across books within the same subject (e.g., ``velocity'' appearing in both 8th- and 9th-grade physics). (v)~We run depth-first cycle detection on the \texttt{is\_a} and \texttt{prerequisites\_for} subgraphs and manually resolve any violations, yielding valid DAGs for the taxonomic and prerequisite relations.

\paragraph{Quality control.} Beyond automated DAG validation, the extraction prompt explicitly discourages hallucination and restricts outputs to ``truly important, clearly presented'' knowledge; edges are encouraged to carry a \texttt{confidence} field and an \texttt{evidence} field linking back to the underlying textbook excerpt; all extracted triples are manually verified by domain experts; and hierarchical deduplication uses light-normalized name matching followed by expert review to reconcile cross-book aliases.

\subsection{Graph Statistics}
\label{sec:stats}

Table~\ref{tab:kg_stats} and Table~\ref{tab:kg_stats_mm} summarize the core node and edge types of K12-KGraph across subjects. For clarity and compactness, we focus on pedagogically salient content nodes (\texttt{Concept}, \texttt{Skill}, \texttt{Experiment}, \texttt{Exercise}, \texttt{Figure}, \texttt{VisualElement}) together with \texttt{Book}, and omit structural container nodes (\texttt{Chapter}, \texttt{Section}), which primarily serve organizational roles. We further omit ubiquitous container relations such as \texttt{appears\_in} and \texttt{is\_part\_of}. By construction, every content node is anchored to a specific textbook location via \texttt{appears\_in}, and all sections and chapters participate in a fixed hierarchy via \texttt{is\_part\_of}; their presence is guaranteed by the schema and therefore carries no information for summarizing graph composition.

\section{Benchmark and Training Data from K12-KGraph}
\label{sec:kg-grounded}

\begin{figure*}[t]
  \centering
  \includegraphics[width=1\linewidth]{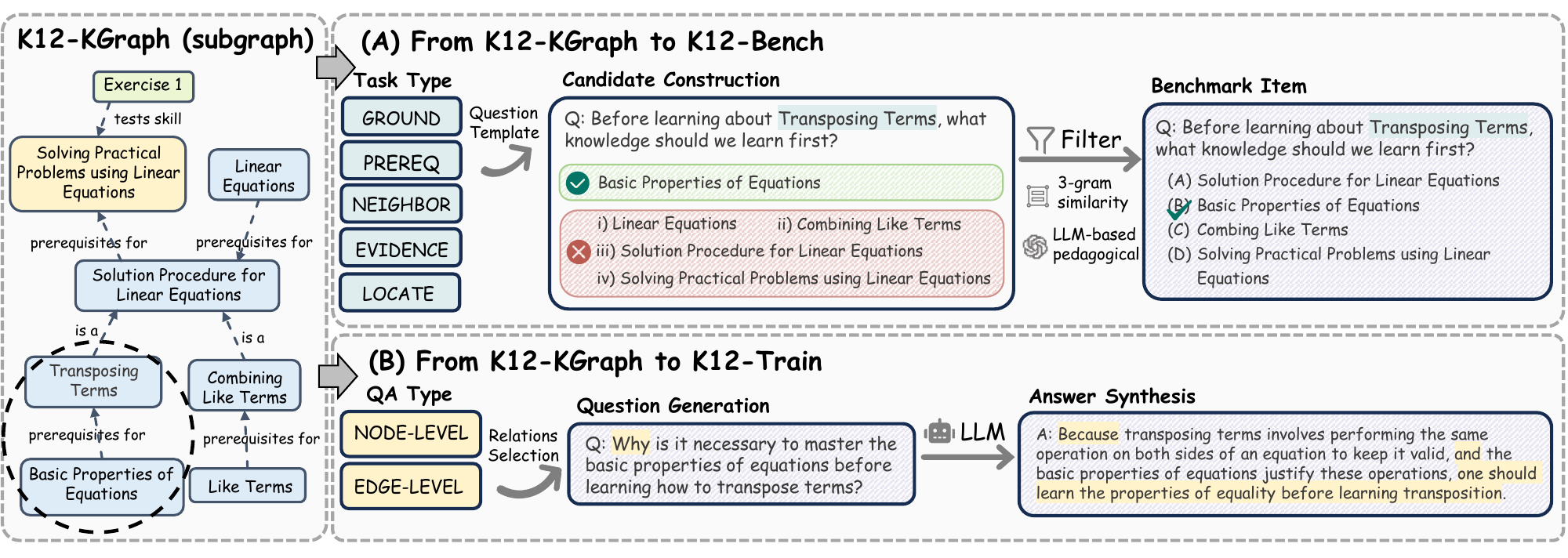}
  \caption{\textbf{A concrete example of K12-Bench and K12-Train generation from K12-KGraph.} (A) shows benchmark construction for a \textsc{Prereq} task via neighborhood sampling and filtering. (B) shows QA synthesis grounded in the \texttt{prerequisites\_for} relation.}
  \label{fig:example}
  \vspace{-0.1in}
\end{figure*}

We derive two KG-grounded datasets from K12-KGraph. K12-Bench (\S\ref{sec:bench}) converts graph textual neighborhoods into multi-select questions for \emph{evaluating} LLMs' curriculum cognition, while K12-Train (\S\ref{sec:train}) converts node attributes and edge semantics into structurally grounded QA pairs for \emph{training} educational LLMs. Figure~\ref{fig:example} illustrates the shared pipeline on a single \texttt{prerequisites\_for} subgraph: panel~(A) shows how the subgraph is instantiated into a \textsc{Prereq} benchmark item, while panel~(B) shows how the same relation is reformulated into a KG-guided QA pair for training. Despite serving different purposes, both resources exploit the same property: every sample can be traced back to a specific subgraph, making difficulty, coverage, and factual correctness systematically controllable.

\subsection{K12-Bench: Benchmark Construction}
\label{sec:bench}

K12-Bench comprises five task families (nine subtasks) that each probe a distinct facet of structural curriculum understanding. All tasks are formulated as multi-select questions: given a question and four labeled candidates, the model must output the full set of correct labels. The gold answer cardinality varies per item from 1 to 3 depending on the query and the local graph neighborhood, so even items with a single correct option must still be produced in the multi-select output format. Questions and distractors are \emph{graph-derived} rather than LLM-generated: for each task we define templates instantiated via graph queries, with correct answers being the true graph neighbors and distractors sampled from a multi-level pool of structurally proximate but non-answer nodes (e.g., 2-hop neighborhoods or siblings under shared \texttt{is\_a} parents). Detailed distractor pool construction and sampling rules are provided in Appendix~\ref{app:distractor}. Models receive only the question text without any graph context, so the benchmark probes \emph{parametric} curriculum knowledge. Each candidate set is filtered by a character 3-gram cosine-similarity step that removes surface-form near-duplicates, followed by a GPT-5.2-based pedagogical filter that discards distractors synonymous with any correct answer (see Appendix~\ref{app:distractor} for details).

We name the five task families \textsc{Ground}, \textsc{Prereq}, \textsc{Neighbor}, \textsc{Evidence}, and \textsc{Locate} after the relation they probe.

\paragraph{\textsc{Ground}: Knowledge Grounding (\texttt{tests\_concept}, \texttt{tests\_skill}).}
\textit{Subtask 1}: given an exercise stem, select the core concepts or skills it tests. \textit{Subtask 2}: given a concept or skill, select which exercises assess it. Distractors are drawn from structurally related concepts/skills within the same curriculum context.

\paragraph{\textsc{Prereq}: Prerequisite Reasoning (\texttt{prerequisites\_for}).}
\textit{Subtask 1}: given a concept/skill, select its prerequisite closure. \textit{Subtask 2}: given a concept/skill, select all of its \emph{most direct} successors. Distractors are sampled from structurally adjacent nodes, including related concepts and taxonomic siblings.

\paragraph{\textsc{Neighbor}: Neighbor Recommendation (\texttt{is\_a}, \texttt{relates\_to}).}
Given a concept, select all \emph{directly} related concepts (via \texttt{is\_a} or \texttt{relates\_to} in either direction). Distractors are drawn from structurally nearby but non-neighbor concepts, primarily from the 2-hop outer ring.

\paragraph{\textsc{Evidence}: Experiment Evidence Chain (\texttt{verifies}).}
\textit{Subtask 1}: given a concept, select which experiments verify it. \textit{Subtask 2}: given an experiment, select which concepts it verifies. Distractors are sampled from structurally related concepts or experiments.

\paragraph{\textsc{Locate}: Cross-Chapter Indexing (\texttt{appears\_in}, \texttt{leads\_to}).}
\textit{Subtask 1}: given a knowledge entity, select the chapter(s) where it \emph{first} appears. \textit{Subtask 2}: given a chapter, select which chapters are its prerequisites. Distractors are drawn from alternative structural locations within the curriculum.

\paragraph{Benchmark statistics and quality.}
K12-Bench contains 23{,}640 multi-select items in total, with per-task breakdown reported in Appendix~\ref{app:bench_composition} (Table~\ref{tab:bench_stats}). Because every item is derived deterministically from validated K12-KGraph subgraphs rather than generated by an LLM, its factual correctness reduces to the correctness of the underlying graph. K12-KGraph itself is fully human-verified by 12 subject-qualified annotators with strong inter-annotator agreement (Fleiss' $\kappa = 0.84$ overall; see Appendix~\ref{app:validation}), and a stratified spot-check of K12-Bench finds $98.4\%$ of sampled items to be fully correct. This pipeline gives K12-Bench an unusually tight link between benchmark quality and graph quality: improvements to the KG propagate directly into the benchmark, while failures are localizable to specific graph edges.

\subsection{K12-Train: KG-Guided Data Synthesis}
\label{sec:train}

K12-Train converts the knowledge graph into supervised fine-tuning data along three complementary paths, illustrated in Appendix~\ref{app:qa_prompts}.

\emph{(i) Node-grounded QA (LLM-prompted).} For each \texttt{Concept}, \texttt{Skill}, \texttt{Experiment}, or \texttt{Exercise} node, we prompt Qwen3-235B-A22B~\citep{qwen2025qwen3} to generate a question--answer pair grounded in that node's typed properties: definitions, formulas, and worked examples for \texttt{Concept}; procedural steps and application scenarios for \texttt{Skill}; instruments, phenomena, and conclusions for \texttt{Experiment}; and the stem augmented with concise step-by-step reasoning for \texttt{Exercise}. This yields supervision that teaches the \emph{content} of a node.

\emph{(ii) Edge-grounded QA (LLM-prompted).} For each semantic relation, the prompt is anchored in a specific question template that forces the answer to articulate the relation itself: \textit{``Why does A belong to category B?''} for \texttt{is\_a}, \textit{``Why must one learn A before B?''} for \texttt{prerequisites\_for}, \textit{``How are A and B related?''} for \texttt{relates\_to}, \textit{``How does experiment E verify concept C?''} for \texttt{verifies}, \textit{``How does figure F explain concept C?''} for \texttt{illustrates}, \textit{``How does a local visual element V correspond to concept C?''} for \texttt{refers\_to}, \textit{``What information does a figure F provide for an exercise?''} for \texttt{requires\_figure}, and \textit{``How does figure F provide visual evidence for an existing relation?''} for \texttt{supports\_edge}. This yields supervision that teaches the \emph{structure} between nodes.

\emph{(iii) Exercise-assessment QA (deterministic templates).} For \texttt{tests\_concept} and \texttt{tests\_skill} edges, which are factually unambiguous, we bypass the LLM and fill deterministic templates directly from the edge. This guarantees full factual grounding and eliminates any risk that the LLM fabricates or paraphrases the target relation for the exercise-to-concept/skill subset.

\paragraph{Quality control.}
To minimize hallucination, we crop each node's property set to only those fields relevant for the target QA type and filter edges with confidence below a specified threshold before prompting, preventing the LLM from drifting into unrelated attributes. Prompts further instruct the model to produce grade-appropriate language, strictly grounded in the provided attributes, with a structured answer format that highlights key conclusions. After generation, every QA pair is validated for JSON structure, non-empty fields, and language consistency. 

The final dataset, \textbf{K12-Train-Full}, contains 7{,}335 QA pairs and is partitioned by modality according to whether answering requires visual input. QA pairs that can be answered from textual information alone constitute \textbf{K12-Train-Text}, while those that require an accompanying figure or visual element constitute \textbf{K12-Train-MM}.

\textbf{K12-Train-Text} contains 2{,}267 QA pairs, selected from a substantially larger KG-derived candidate pool through source-balanced random subsampling. 
Specifically, it includes 450 \texttt{Exercise} examples (with step-by-step reasoning), 356 \texttt{tests\_concept}/\texttt{tests\_skill} examples, 695 node-grounded QA pairs (from \texttt{Concept}, \texttt{Skill}, and \texttt{Experiment}), and 766 relation-grounded QA pairs (from \texttt{is\_a}, \texttt{prerequisites\_for}, \texttt{relates\_to}, and \texttt{verifies}). 
\textbf{K12-Train-MM} contains 5{,}068 QA pairs, all of which are relation-grounded QA pairs (\texttt{illustrates}, \texttt{refers\_to}, \texttt{requires\_figure}, and \texttt{supports\_edge}) selected based on the confidence of corresponding edges ($1.0$ for \texttt{illustrates}, \texttt{refers\_to}, \texttt{supports\_edge}, and $0.8$ for \texttt{requires\_figure}) to meet the corresponding target size. 
This balancing controls data quality while mitigating the natural skew in the raw KG-derived pool (e.g., relatively fewer textbook exercises), and encourages the model to learn uniformly across content and relational supervision signals. 
Despite its modest size, we show in Section~\ref{sec:experiments} that these structurally grounded samples are highly effective for educational SFT.
% For multimodal QA, we retain all \texttt{requires\_figure} relations with confidence at least 0.8 and sample confidence-1.0 \texttt{supports\_edge}, \texttt{illustrates}, and \texttt{refers\_to} relations by book and source section; \texttt{VisualElement}-based QA additionally requires bounding-box confidence of at least 0.9.

\begin{table*}[t]
\centering
\caption{\textbf{K12-Bench results (zero-shot, answer-only), reported in \%.} EM = exact match (per-instance all-or-nothing, averaged across instances). F1 is the instance-level (example-based) macro F1: we compute precision/recall/F1 on each instance's option-label set, then average across instances. Within each model family (open-source vs.\ proprietary), the best Overall score is highlighted in yellow, and the second-best in green.}
\label{tab:bench_results}
\resizebox{\linewidth}{!}{
\begin{tabular}{l cc cc cc cc cc cc}
\toprule
\multirow{2}{*}{\textbf{Model}} & \multicolumn{2}{c}{\textsc{Ground}} & \multicolumn{2}{c}{\textsc{Prereq}}
& \multicolumn{2}{c}{\textsc{Neighbor}} & \multicolumn{2}{c}{\textsc{Evidence}} & \multicolumn{2}{c}{\textsc{Locate}} & \multicolumn{2}{c}{\textbf{Overall}} \\
\cmidrule(lr){2-3} \cmidrule(lr){4-5} \cmidrule(lr){6-7} \cmidrule(lr){8-9} \cmidrule(lr){10-11} \cmidrule(lr){12-13}
& \textbf{EM} & \textbf{F1} & \textbf{EM} & \textbf{F1} & \textbf{EM} & \textbf{F1} & \textbf{EM} & \textbf{F1} & \textbf{EM} & \textbf{F1} & \textbf{EM} & \textbf{F1} \\
\midrule
\multicolumn{13}{c}{\cellcolor{model_type}\textbf{Random Baseline}} \\
\midrule
    \textit{Random guess}
    & 6.7 & 36.2 & 6.7 & 37.9 & 6.7 & 41.3 & 6.7 & 37.7 & 6.7 & 32.9 & 6.7 & 36.4 \\
\midrule
\multicolumn{13}{c}{\cellcolor{model_type}\textbf{Open Source Models}} \\
\midrule
    \textit{Meta-LLaMA-3-8B-Instruct}
    & 6.2 & 54.9 & 4.3 & 47.9 & 3.8 & 53.4 & 5.2 & 55.4 & 11.5 & 53.9 & 7.2 & 52.6 \\
    \textit{GLM-4.7-Flash}
    & 36.4 & 70.9 & 13.4 & 56.6 & 15.2 & 59.6 & 39.3 & 72.8 & 48.9 & 66.3 & 31.7 & 63.9 \\
    \textit{Ministral-3-14B-Instruct}
    & 43.0 & 75.1 & 18.4 & 59.4 & 14.5 & 59.8 & 38.3 & 73.2 & 59.3 & 72.2 & 37.5 & \cellcolor{column_green}67.4 \\
    \textit{Qwen3-32B}
    & 46.7 & 77.2 & 16.9 & 60.4 & 14.6 & 60.3 & 41.7 & 75.1 & 72.1 & 75.9 & \cellcolor{column_green}42.6 & \cellcolor{column_yellow}69.5 \\
    \textit{Gemma-4-31B-IT}
    & 50.6 & 79.0 & 28.3 & 62.6 & 15.0 & 60.7 & 43.3 & 73.9 & 73.4 & 73.5 & \cellcolor{column_yellow}46.4 & \cellcolor{column_yellow}69.5 \\
\midrule
\multicolumn{13}{c}{\cellcolor{model_type}\textbf{Proprietary Models}} \\
\midrule
    \textit{GPT-4o}
    & 30.3 & 70.6 & 9.3 & 57.5 & 10.1 & 57.9 & 32.4 & 71.8 & 55.6 & 72.3 & 31.1 & 65.9 \\
    \textit{GPT-5-mini}
    & 30.4 & 70.4 & 9.9 & 57.6 & 12.5 & 59.0 & 32.9 & 72.3 & 55.5 & 72.8 & 31.7 & 66.4 \\
    \textit{GPT-5.2}
    & 50.8 & 79.2 & 17.9 & 59.5 & 13.1 & 60.0 & 41.6 & 73.9 & 70.1 & 71.5 & 42.8 & \cellcolor{column_green}68.0 \\
    \textit{Gemini-2.5-Flash}
    & 58.0 & 75.7 & 29.7 & 56.3 & 15.4 & 56.0 & 47.1 & 72.0 & 72.8 & 74.0 & \cellcolor{column_green}48.3 & 66.7 \\
    \textit{Gemini-3-Flash}
    & 63.4 & 83.3 & 34.8 & 58.2 & 33.4 & 63.5 & 47.4 & 72.4 & 81.7 & 82.6 & \cellcolor{column_yellow}57.1 & \cellcolor{column_yellow}73.0 \\
\bottomrule
\end{tabular}}
\vspace{-2mm}
\end{table*}

\section{Experiments and Results}
\label{sec:experiments}

We conduct two complementary experiments:
(i)~\textbf{benchmarking} open-source and proprietary LLMs on K12-Bench to assess structural curriculum understanding (\S\ref{sec:exp_bench}), and
(ii)~\textbf{fine-tuning} both LLMs and VLMs on K12-Train and on other mainstream instruction-tuning datasets, and comparing their performance on downstream educational tasks to validate the validity of our data (\S\ref{sec:exp_sft}).

\begin{table*}[t]
\centering
\caption{\textbf{GaokaoBench scores after SFT on Qwen3-4B-Base and Llama3.1-8B-Base with 2{,}300 samples from each dataset.} ``Total'' denotes the sum of per-subject scores; ``Obj'' / ``Sub'' are overall score rates across objective / subjective sub-portions. Within each base-model block, the best Overall score per column is highlighted in yellow, and the second-best in green.}
\label{tab:gaokao-bench}
\small
\setlength{\tabcolsep}{4pt}
\resizebox{\linewidth}{!}{
\begin{tabular}{l rrrrrrrrrr rrr}
\toprule
\multirow{2}{*}{\textbf{Model}} & \multirow{2}{*}{\textbf{Eng.}} & \multirow{2}{*}{\textbf{Chi.}}
& \textbf{Sci.} & \textbf{Hum.}
& \multirow{2}{*}{\textbf{Phy.}} & \multirow{2}{*}{\textbf{Chem.}} & \multirow{2}{*}{\textbf{Bio.}} & \multirow{2}{*}{\textbf{His.}} & \multirow{2}{*}{\textbf{Geo.}} & \multirow{2}{*}{\textbf{Pol.}} & \multicolumn{3}{c}{\textbf{Overall}} \\
\cmidrule(lr){12-14}
& & & \textbf{Math} & \textbf{Math} & & & & & & & \textbf{Total} & \textbf{Obj} & \textbf{Sub} \\
\midrule
\multicolumn{14}{c}{\cellcolor{model_type}\textbf{Trained on Qwen3-4B-Base}} \\
\midrule
    \textit{Qwen3-4B-Base}
    & 37.88 & 55.20 & 65.22 & 76.59 & 47.34 & 30.60 & 30.72 & 39.80 & 17.72 & 44.35 & 445.42 & 61.4\% & 19.9\% \\
    \textit{Qwen3-4B-Instruct}
    & 119.98 & 103.22 & 97.32 & 106.62 & 72.60 & 54.45 & 73.40 & 88.75 & 88.38 & 90.65 & 895.37 & 77.7\% & 77.3\% \\
    \midrule
    + OpenHermes
    & 125.04 & 114.28 & 117.87 & 123.75 & 69.96 & 69.85 & 80.11 & 87.65 & 86.18 & 92.15 & 966.84 & 79.9\% & 85.4\% \\
    + Infinity
    & 122.92 & 103.74 & 119.55 & 120.90 & 73.08 & 73.45 & 77.97 & 82.95 & 86.42 & 92.75 & 953.73 & 80.1\% & 82.4\% \\
    + UltraChat
    & 120.44 & 110.19 & 100.32 & 104.58 & 69.17 & 66.00 & 77.11 & 81.00 & 82.78 & 91.05 & 902.64 & 74.6\% & 79.8\% \\
    + WizardLM
    & 124.28 & 112.52 & 102.69 & 112.26 & 69.17 & 65.70 & 78.94 & 81.95 & 83.44 & 91.80 & 922.75 & 77.6\% & 80.6\% \\
    + DataFlow
    & 120.93 & 116.85 & 122.49 & 127.92 & 77.35 & 70.70 & 81.79 & 87.65 & 85.78 & 94.45 & \cellcolor{column_green}985.91 & \cellcolor{column_green}81.3\% & \cellcolor{column_green}87.1\% \\
    + LMSYS
    & 86.39 & 85.16 & 80.58 & 89.70 & 57.40 & 60.80 & 73.07 & 78.55 & 83.86 & 86.85 & 782.36 & 65.1\% & 68.9\% \\
    + SmolTalk
    & 119.72 & 111.69 & 117.93 & 124.92 & 77.09 & 68.75 & 79.07 & 85.05 & 85.18 & 94.20 & 963.60 & 78.9\% & 85.4\% \\
    + Tulu
    & 122.89 & 93.58 & 117.60 & 125.91 & 74.12 & 74.15 & 81.25 & 84.35 & 84.50 & 93.30 & 951.65 & 80.7\% & 81.4\% \\
    \midrule
    \textbf{+ K12-Train-Text (ours)}
    & 122.04 & 120.18 & 128.61 & 132.00 & 79.73 & 79.35 & 80.74 & 87.40 & 85.76 & 94.15 & \cellcolor{column_yellow}1009.96 & \cellcolor{column_yellow}81.8\% & \cellcolor{column_yellow}89.5\% \\
\midrule
\multicolumn{14}{c}{\cellcolor{model_type}\textbf{Trained on Llama3.1-8B-Base}} \\
\midrule
    \textit{Llama3.1-8B-Base}
    & 26.89 & 16.83 & 6.60 & 6.48 & 16.19 & 8.05 & 8.69 & 18.50 & 17.18 & 5.35 & 130.76 & 12.7\% & 9.0\% \\
    \textit{Llama3.1-8B-Instruct}
    & 14.49 & 28.52 & 58.35 & 57.90 & 40.94 & 23.85 & 33.14 & 52.05 & 43.92 & 51.35 & 404.50 & 33.0\% & 36.2\% \\
    \midrule
    + OpenHermes
    & 62.88 & 79.84 & 51.54 & 54.39 & 38.83 & 39.45 & 56.93 & 65.60 & 65.92 & 64.80 & 580.19 & 35.7\% & 62.4\% \\
    + Infinity
    & 67.50 & 79.75 & 48.57 & 52.38 & 37.38 & 41.40 & 53.13 & 67.05 & 61.12 & 65.05 & 573.33 & 37.1\% & 60.5\% \\
    + UltraChat
    & 72.97 & 83.17 & 52.02 & 56.31 & 32.54 & 52.40 & 56.87 & 64.20 & 58.24 & 63.50 & 592.23 & 37.3\% & 62.7\% \\
    + WizardLM
    & 89.76 & 79.78 & 51.24 & 52.26 & 33.59 & 46.60 & 54.36 & 66.40 & 55.58 & 63.50 & \cellcolor{column_green}593.08 & \cellcolor{column_green}39.2\% & 62.8\% \\
    + DataFlow
    & 62.46 & 86.82 & 47.97 & 50.91 & 33.18 & 40.45 & 57.29 & 70.20 & 67.60 & 68.15 & 585.03 & 32.6\% & \cellcolor{column_yellow}66.8\% \\
    + LMSYS
    & 60.36 & 40.16 & 34.32 & 43.56 & 24.20 & 28.70 & 36.90 & 48.25 & 48.22 & 37.95 & 402.62 & 29.1\% & 38.5\% \\
    + SmolTalk
    & 62.02 & 80.40 & 44.67 & 51.87 & 31.86 & 34.90 & 53.45 & 67.90 & 65.02 & 63.55 & 555.64 & 34.7\% & 60.8\% \\
    + Tulu
    & 74.73 & 74.53 & 56.58 & 60.15 & 36.76 & 40.05 & 55.53 & 63.40 & 62.68 & 62.55 & 586.97 & 38.8\% & 61.1\% \\
    \midrule
    \textbf{+ K12-Train-Text (ours)}
    & 76.17 & 93.39 & 52.02 & 60.45 & 34.63 & 48.85 & 59.61 & 69.80 & 67.62 & 62.95 & \cellcolor{column_yellow}625.49 & \cellcolor{column_yellow}41.0\% & \cellcolor{column_green}65.2\% \\
\bottomrule
\end{tabular}}
\vspace{-2mm}
\end{table*}

\subsection{Experimental Settings}
\label{sec:exp_settings}
\paragraph{Models under evaluation.} For K12-Bench, we evaluate five open-source LLMs (Meta-LLaMA-3-8B-Instruct~\citep{grattafiori2024llama3}, GLM-4.7-Flash~\citep{glm2024chatglm}, Ministral-3-14B-Instruct~\citep{liu2026ministral}, Qwen3-32B~\citep{qwen2025qwen3}, Gemma-4-31B-IT~\citep{team2024gemma}) and five proprietary APIs (GPT-4o, GPT-5-mini, GPT-5.2, Gemini-2.5-Flash, Gemini-3-Flash) in a zero-shot, answer-only setting: given a question and four labeled options, the model outputs the correct option label(s). We report per-task exact match (EM; all predicted labels must equal the gold set) and an instance-level (example-based) \emph{macro F1}: for each instance we compute precision, recall, and F1 on the predicted and gold option-label sets, and then average across instances, so every question contributes equally regardless of how many labels it has. This choice keeps F1 on the same per-instance footing as EM, avoids letting items with larger gold sets dominate a micro-pooled F1, and matches the \texttt{average=`samples'} convention for multi-label classification. Task-level scores are the average over instances within that task, and Overall scores are the instance-count-weighted average across tasks.

\paragraph{Fine-tuning settings.}
Our fine-tuning experiments consist of two evaluation settings. 
For \textbf{text-only} evaluation, we train Qwen3-4B-Base and Llama3.1-8B-Base on K12-Train-Text and eight general instruction-tuning datasets, with all datasets matched to a budget of approximately 2{,}300 samples. 
For \textbf{multimodal} evaluation, we fine-tune Qwen3.5-2B-Base on K12-Train-Full, K12-Train-Text, and K12-Train-MM, together with the full available DataFlow and WizardLM datasets (10{,}000 and 142{,}759 samples, respectively).

\paragraph{Text-only SFT protocol.}
We perform full-parameter SFT on Qwen3-4B-Base~\citep{qwen2025qwen3} and Llama3.1-8B-Base~\citep{grattafiori2024llama} using K12-Train-Text and eight general-purpose instruction-tuning datasets: OpenHermes-2.5~\citep{OpenHermes25}, Infinity-Instruct~\citep{li2025infinity}, UltraChat~\citep{ding2023enhancing}, WizardLM\_evol\_instruct\_V2\_196k~\citep{luo2023wizardcoder}, DataFlow-10K-Instruct~\citep{liang2025dataflow}, LMSYS~\citep{zheng2023lmsys}, SmolTalk~\citep{allal2025smollm2}, and Tulu-3-SFT~\citep{lambert2024tulu}. To ensure a fair comparison, we uniformly sample 2{,}300 examples from each baseline, approximately matching the 2{,}267 examples in K12-Train-Text. All configurations use identical hyperparameters: learning rate $5\times10^{-6}$, cosine scheduler with 10\% warmup, 3 epochs, per-device batch size 1 with gradient accumulation of 4, DeepSpeed ZeRO-3, and bf16 precision, implemented via LLaMA-Factory~\citep{zheng2024llamafactory}. For each backbone, we additionally include two reference points that are \emph{not} constrained to 2{,}300 samples: the untuned base and the corresponding official instruct version.

\paragraph{Multimodal SFT protocol.}
We perform LoRA SFT on Qwen3.5-2B-Base using K12-Train-Text, K12-Train-MM, and K12-Train-Full, as well as the full DataFlow and WizardLM datasets containing 10{,}000 and 142{,}759 examples, respectively. All configurations use identical hyperparameters: learning rate $1\times10^{-4}$, cosine scheduler with 10\% warmup, 3 epochs, per-device batch size 4 with gradient accumulation of 2, and bf16 precision, implemented via LLaMA-Factory~\citep{zheng2024llamafactory}. We also additionally report the untuned Qwen3.5-2B-Base model and its official instruction-tuned variant as reference points.

\paragraph{Evaluation axes.} Text-only fine-tuned models are evaluated on two open-ended educational benchmarks: \textbf{GaokaoBench}~\citep{zhang2023gaokao} and \textbf{EduEval}~\citep{ma2025edueval}. For GaokaoBench, we follow the official evaluation protocol for both objective and subjective questions, using string-matching against the reference answers. For EduEval, we report a subset of 18 tasks covering five capability dimensions (Application, Ethics, Memory, Reasoning, and Understanding). As EduEval employs heterogeneous evaluation protocols across tasks, we restrict to those with standardized, officially supported scoring to ensure comparability. Multimodal fine-tuned models are evaluated on three complementary benchmarks. \textbf{Gaokao-MM}~\citep{zong2024gaokao} consists of image-bearing multiple-choice questions from eight Gaokao subjects; following its official protocol, we apply rule-based answer extraction and report accuracy. For \textbf{MDK12-Bench}~\citep{zhou2026mdk12}, which covers six disciplines and five question formats, we use the official judge-assisted evaluation pipeline and aggregate scores across question formats and disciplines. \textbf{K12Vista}~\citep{li2025k12vista} covers five subjects  with multiple-choice, fill-in-the-blank, and open-ended questions; following its official protocol, we use the K12-PEM judge to assess answer correctness and reasoning quality, and report per-format mean scores and an overall mean score.

\subsection{Benchmarking LLMs on K12-Bench}
\label{sec:exp_bench}

Table~\ref{tab:bench_results} presents per-task and overall results across all five K12-Bench task families. In addition to model results, we report a simple \textit{Random guess} baseline, which samples uniformly from the 15 non-empty label subsets of $\{A,B,C,D\}$. This baseline provides a lower reference for purely random guessing on multi-select items; formal EM/F1 definitions and the expectation formulas are in Appendix~\ref{app:baseline_metrics}. 

\begin{table*}[t]
\centering
\caption{\textbf{EduEval scores after SFT on Qwen3-4B-Base and Llama3.1-8B-Base, organized by capability dimension (18 sub-tasks).} Within each base-model block, the best Avg is highlighted in yellow, and the second-best in green.}
% ``Avg'' denotes the weighted mean over all sub-tasks. For ES (Essay Scoring), where the original metric is RMSE (lower is better), we report $100 - \mathrm{RMSE}$ so that all metrics are aligned on a higher-is-better scale.
\label{tab:edueval}
\setlength{\tabcolsep}{3pt}
\resizebox{\linewidth}{!}{
\begin{tabular}{l ccccc cccc ccc ccc ccc c}
\toprule
\multirow{2}{*}{\textbf{Model}}
& \multicolumn{5}{c}{\textbf{Application}}
& \multicolumn{4}{c}{\textbf{Ethics}}
& \multicolumn{3}{c}{\textbf{Memory}}
& \multicolumn{3}{c}{\textbf{Reasoning}}
& \multicolumn{3}{c}{\textbf{Understanding}} & \multirow{2}{*}{\textbf{Avg.}}\\
\cmidrule(lr){2-6} \cmidrule(lr){7-10} \cmidrule(lr){11-13} \cmidrule(lr){14-16} \cmidrule(lr){17-19}
& \textbf{CDC} & \textbf{ES} & \textbf{JPS} & \textbf{PPS} & \textbf{SPS}
& \textbf{EEJ} & \textbf{JES} & \textbf{PM} & \textbf{SES}
& \textbf{JKR} & \textbf{PFR} & \textbf{SCR}
& \textbf{JR} & \textbf{PR} & \textbf{SR}
& \textbf{JU} & \textbf{PU} & \textbf{SU} \\
\midrule
\multicolumn{20}{c}{\cellcolor{model_type}\textbf{Trained on Qwen3-4B-Base}} \\
   \midrule
    \textit{Qwen3-4B-Base}
    & 15.7 & 25.6 & 35.2 & 33.0 & 32.2
    & 41.0 & 44.0 & 42.4 & 39.4
    & 34.0 & 35.6 & 51.4
    & 33.2 & 29.8 & 34.4
    & 34.1 & 28.9 & 42.9 & 34.51 \\
    \textit{Qwen3-4B-Instruct}
    & 19.2 & 90.1 & 74.0 & 58.0 & 63.0
    & 65.8 & 72.4 & 67.6 & 64.6
    & 63.5 & 73.8 & 75.8
    & 75.4 & 54.8 & 70.0
    & 78.4 & 78.3 & 78.6 & 66.30 \\
    \midrule
    + OpenHermes
    & 14.8 & 89.8 & 74.6 & 59.8 & 61.4
    & 65.0 & 70.8 & 68.8 & 68.2
    & 63.0 & 70.4 & 75.8
    & 76.2 & 56.4 & 69.8
    & 78.5 & 77.4 & 80.1 & 66.01 \\
    + Infinity
    & 15.9 & 91.3 & 73.4 & 56.4 & 62.6
    & 66.6 & 72.2 & 67.2 & 68.4
    & 62.5 & 72.0 & 77.4
    & 79.0 & 56.4 & 71.4
    & 76.8 & 76.1 & 79.3 & 66.04 \\
    + UltraChat
    & 16.4 & 92.0 & 74.4 & 52.2 & 59.2
    & 64.8 & 74.2 & 70.0 & 67.6
    & 64.0 & 71.4 & 80.4
    & 74.8 & 52.4 & 68.4
    & 76.4 & 76.2 & 79.6 & 65.44 \\
    + WizardLM
    & 16.4 & 91.1 & 74.8 & 59.6 & 64.4
    & 69.2 & 74.8 & 71.6 & 71.4
    & 66.8 & 70.2 & 78.2
    & 75.0 & 52.0 & 70.0
    & 76.6 & 76.1 & 80.9 & \cellcolor{column_green}66.70 \\
    + DataFlow
    & 13.7 & 92.1 & 72.2 & 57.2 & 61.0
    & 67.6 & 74.8 & 72.0 & 70.4
    & 66.8 & 69.6 & 74.2
    & 76.6 & 51.4 & 67.6
    & 77.2 & 78.1 & 77.8 & 65.57 \\
    + LMSYS
    & 14.2 & 50.2 & 72.0 & 57.0 & 64.4
    & 68.8 & 72.8 & 70.4 & 69.0
    & 63.5 & 71.0 & 75.4
    & 75.0 & 53.8 & 67.4
    & 73.4 & 76.5 & 77.0 & 64.68 \\
    + SmolTalk
    & 14.6 & 92.7 & 73.4 & 58.8 & 65.4
    & 69.0 & 74.8 & 70.4 & 70.6
    & 63.7 & 71.4 & 76.0
    & 76.0 & 52.6 & 67.6
    & 75.1 & 77.6 & 80.0 & 66.10 \\
    + Tulu
    & 15.8 & 93.2 & 75.4 & 59.0 & 64.2
    & 69.2 & 74.0 & 70.8 & 69.4
    & 63.0 & 71.0 & 79.0
    & 75.6 & 52.8 & 67.0
    & 76.2 & 77.2 & 79.5 & 66.27 \\
    \midrule
    \textbf{+ K12-Train-Text (ours)}
    & 16.2 & 93.1 & 74.4 & 59.6 & 60.2
    & 71.8 & 77.0 & 73.6 & 71.6
    & 69.2 & 70.6 & 78.4
    & 72.6 & 57.8 & 68.2
    & 74.8 & 77.8 & 79.0 & \cellcolor{column_yellow}66.76 \\
\midrule
\multicolumn{20}{c}{\cellcolor{model_type}\textbf{Trained on Llama3.1-8B-Base}} \\
\midrule
    \textit{Llama3.1-8B-Base}
    & 10.7 & 55.8 & 24.6 & 13.2 & 21.4
    & 25.4 & 26.4 & 26.0 & 23.4
    & 15.2 & 25.2 & 40.0
    & 11.2 & 12.0 & 19.6
    & 25.0 & 10.2 & 21.9 & 20.98 \\
    \textit{Llama3.1-8B-Instruct}
    & 10.7 & 83.4 & 22.6 & 26.8 & 25.8
    & 34.0 & 32.8 & 34.0 & 35.6
    & 33.2 & 22.2 & 24.4
    & 27.4 & 26.6 & 24.0
    & 28.1 & 31.5 & 28.4 & 27.31 \\
    \midrule
    + OpenHermes
    & 12.7 & 80.7 & 42.4 & 27.8 & 34.4
    & 55.6 & 58.6 & 60.0 & 60.6
    & 34.8 & 34.4 & 37.8
    & 40.8 & 29.0 & 38.4
    & 39.1 & 43.7 & 41.3 & \cellcolor{column_green}40.05 \\
    + Infinity
    & 16.0 & 89.6 & 38.8 & 23.0 & 32.0
    & 49.4 & 56.2 & 52.8 & 57.2
    & 27.3 & 40.2 & 41.2
    & 38.4 & 29.0 & 38.4
    & 37.1 & 32.3 & 41.3 & 37.33 \\
    + UltraChat
    & 12.9 & 89.1 & 34.8 & 36.8 & 25.6
    & 61.6 & 34.6 & 33.2 & 34.6
    & 25.5 & 33.4 & 22.8
    & 24.8 & 27.6 & 26.8
    & 39.9 & 34.0 & 30.8 & 31.87 \\
    + WizardLM
    & 16.4 & 88.6 & 41.8 & 21.8 & 33.0
    & 56.6 & 62.2 & 61.2 & 63.8
    & 25.8 & 36.0 & 39.6
    & 39.4 & 26.2 & 33.0
    & 39.8 & 32.6 & 38.7 & 38.58 \\
    + DataFlow
    & 9.6 & 80.4 & 33.0 & 25.8 & 28.4
    & 49.4 & 48.4 & 57.2 & 53.6
    & 34.0 & 31.0 & 29.0
    & 35.6 & 30.6 & 29.8
    & 31.9 & 37.1 & 33.8 & 34.53 \\
    + LMSYS
    & 21.7 & 74.2 & 27.0 & 29.6 & 27.0
    & 42.0 & 42.2 & 40.8 & 40.8
    & 24.8 & 22.8 & 24.2
    & 24.8 & 26.0 & 27.8
    & 28.7 & 27.9 & 26.3 & 29.85 \\
    + SmolTalk
    & 12.3 & 60.7 & 37.0 & 28.8 & 29.6
    & 60.4 & 66.0 & 64.8 & 65.8
    & 35.8 & 34.2 & 40.6
    & 29.2 & 28.8 & 30.6
    & 33.0 & 36.6 & 36.4 & 37.64 \\
    + Tulu
    & 9.6 & 78.7 & 36.8 & 26.4 & 30.0
    & 53.0 & 58.8 & 58.8 & 56.2
    & 36.0 & 39.4 & 31.2
    & 43.8 & 25.2 & 38.8
    & 39.3 & 41.2 & 39.3 & 37.68 \\
    \midrule
    \textbf{+ K12-Train-Text (ours)}
    & 12.3 & 88.3 & 40.6 & 27.8 & 32.6
    & 60.0 & 67.2 & 69.6 & 68.6
    & 34.8 & 39.8 & 35.0
    & 40.8 & 27.6 & 35.8
    & 40.7 & 41.4 & 41.9 & \cellcolor{column_yellow}40.90 \\
\bottomrule
\end{tabular}}
\vspace{-2mm}
\end{table*}

\paragraph{Findings.}
\textbf{(1) Even strong LLMs struggle with curriculum structure.} The best overall exact match is only 57.1\% (Gemini-3-Flash); a strong open-source model (Gemma-4-31B-IT) reaches just 46.4\%. In more than half the questions, models fail to identify the complete correct answer set. Overall F1 peaks at $\sim$73\%, indicating systematic errors rather than isolated mistakes. Moreover, LLaMA-3-8B-Instruct (EM $=7.2\%$) is essentially indistinguishable from pure random guessing (EM $=6.7\%$), showing that smaller open-source models have no parametric knowledge of curriculum structure at all.

\textbf{(2) Prerequisite and neighbor tasks are hardest.} \textsc{Prereq} and \textsc{Neighbor} yield the lowest EM across almost every model, with EM below 35\% even for Gemini-3-Flash on both tasks. These tasks require understanding \emph{directed} and \emph{structural} relationships between concepts, capabilities that current LLMs evidently lack.

\textbf{(3) \textsc{Ground} and \textsc{Evidence} are relatively easier.} Knowledge Grounding (\textsc{Ground}) and Experiment Evidence Chain (\textsc{Evidence}) achieve the highest F1 (above 75\% on \textsc{Ground} and above 72\% on \textsc{Evidence} for top models), likely because exercise--concept and experiment--concept associations are more explicitly discussed in pretraining corpora such as textbook solutions and lab handbooks.

\subsection{K12-Train for Educational SFT}
\label{sec:exp_sft}
% Our fine-tuning experiments consist of two evaluation tracks. 
% For \textit{text-only} evaluation, we train Qwen3-4B-Base and Llama3.1-8B-Base on K12-Train-Text and eight general instruction-tuning datasets, with all SFT datasets matched to a budget of approximately 2{,}300 samples. 
% For \textit{multimodal} evaluation, we fine-tune Qwen3.5-2B-Base on K12-Train-Full, K12-Train-Text, and K12-Train-MM, together with the full available DataFlow and WizardLM datasets (10{,}000 and 142{,}759 samples, respectively).
We first report the text-only fine-tuning results on GaokaoBench and EduEval, followed by the multimodal results on Gaokao-MM, MDK12-Bench, and K12Vista, and then analyze the sources of K12-Train's effectiveness.

\subsubsection{Text-Only Fine-Tuning Results}
\label{sec:exp_sft_text}
\paragraph{GaokaoBench.} Table~\ref{tab:gaokao-bench} reports per-subject scores, the summed total across ten subjects, and the objective/subjective score rates for both backbones. On Qwen3-4B-Base, K12-Train-Text achieves the highest overall total (\textbf{1009.96}), improving over the strongest baseline DataFlow (985.91) by $+24.1$, while also reaching the best objective score rate (81.8\%) and subjective score rate (89.5\%). On Llama3.1-8B-Base, K12-Train-Text again obtains the highest total (\textbf{625.49}) and the best objective score rate (41.0\%), outperforming the strongest baseline WizardLM (593.08) by $+32.4$. Curriculum-structured training thus transfers across distinct model families rather than being confined to a single backbone.

\paragraph{EduEval.} Table~\ref{tab:edueval} presents the 18 sub-task scores grouped into five capability dimensions (Application, Ethics, Memory, Reasoning, Understanding) for both backbones. On Qwen3-4B-Base, K12-Train-Text achieves the best average score (\textbf{66.76}) among all 2{,}300-sample SFT configurations, surpassing the strongest baseline WizardLM (66.70) while delivering the strongest Ethics results overall. On Llama3.1-8B-Base, K12-Train-Text again reaches the best average score (\textbf{40.90}), ahead of OpenHermes (40.05) by nearly one point, a non-trivial gap given identical training budget. Although the overall gains are smaller than on GaokaoBench, they remain consistent across both base models under the same 2{,}300-sample budget, demonstrating robust improvements across backbones.

\subsubsection{Multimodal Fine-Tuning Results}
\label{sec:exp_sft_mm}

\begin{table*}[t]
\centering
\caption{\textbf{Gaokao-MM accuracy (\%) after SFT on Qwen3.5-2B-Base with full samples from each dataset.}}
\label{tab:gaokao-mm}
\small
\setlength{\tabcolsep}{4pt}
\resizebox{\linewidth}{!}{
\begin{tabular}{l cccccccc c}
\toprule
\textbf{Model} & \textbf{Math} & \textbf{Chinese}
& \textbf{Physics} & \textbf{Chemistry} & \textbf{Biology} & \textbf{History} & \textbf{Geography} & \textbf{Politics} & \textbf{Overall} \\
\midrule
\multicolumn{10}{c}{\cellcolor{model_type}\textbf{Trained on Qwen3.5-2B-Base}} \\
\midrule
    \textit{Qwen3.5-2B-Base}
    & 23.7 & 12.5 & 4.6 & 52.2 & 28.6 & 64.7 & 45.2 & 51.5 & 32.4 \\
    \textit{Qwen3.5-2B (Instruct)}
    & 32.5 & 12.5 & 9.8 & 53.7 & 52.4 & 64.7 & 46.6 & 48.5 & 36.1 \\
\midrule
    + WizardLM (Full)
    & 38.8 & 28.1 & 16.7 & 38.8 & 42.9 & 58.8 & 55.2 & 48.5 & \cellcolor{column_green}39.1 \\
    + DataFlow (Full)
    & 37.5 & 18.8 & 19.0 & 47.8 & 23.8 & 64.7 & 34.8 & 39.4 & 33.3 \\
\midrule
    + K12-Train-Text
    & 30.0 & 31.2 & 20.4 & 44.8 & 38.1 & 67.6 & 47.5 & 51.5 & 38.3 \\
    + K12-Train-MM
    & 38.8 & 28.1 & 16.7 & 49.3 & 42.9 & 64.7 & 49.3 & 39.4 & 38.8 \\
    \textbf{+ K12-Train-Full}
    & 28.7 & 25.0 & 18.1 & 46.3 & 52.4 & 67.6 & 50.7 & 51.5 & \cellcolor{column_yellow}39.9 \\
\bottomrule
\end{tabular}}
\end{table*}

\begin{table*}[t]
\centering
\caption{\textbf{MDK12-medium scores after SFT on Qwen3.5-2B-Base with full samples from each dataset.}}
\label{tab:mdk12}
\small
\setlength{\tabcolsep}{4pt}
\resizebox{\linewidth}{!}{
\begin{tabular}{l cccccc c}
\toprule
\textbf{Model} & \textbf{Math} & \textbf{Physics} & \textbf{Chemistry} & \textbf{Biology} & \textbf{Geography} & \textbf{Information Science} & \textbf{Overall} \\
\midrule
\multicolumn{8}{c}{\cellcolor{model_type}\textbf{Trained on Qwen3.5-2B-Base}} \\
\midrule
\textit{Qwen3.5-2B-Base}
& 45.83 & 49.56 & 48.52 & 54.75 & 54.86 & 56.63 & 50.77 \\
\textit{Qwen3.5-2B (Instruct)}
& 48.57 & 47.99 & 46.69 & 57.70 & 55.41 & 50.67 & 50.61 \\
\midrule
+ WizardLM (Full)
& 45.40 & 48.42 & 51.27 & 53.11 & 54.24 & 58.65 & 50.76 \\
+ DataFlow (Full)
& 52.92 & 47.19 & 50.04 & 53.37 & 59.70 & 44.88 & 51.28 \\
\midrule
+ K12-Train-Text
& 49.68 & 51.37 & 52.77 & 53.90 & 53.37 & 50.19 & 51.55 \\
+ K12-Train-MM
& 47.47 & 49.57 & 53.59 & 56.65 & 54.48 & 58.60 & \cellcolor{column_green}52.33 \\
\textbf{+ K12-Train-Full}
& 49.84 & 53.24 & 51.37 & 57.92 & 54.55 & 53.68 & \cellcolor{column_yellow}52.94 \\
\bottomrule
\end{tabular}}
\end{table*}

\begin{table*}[t]
\centering
\caption{\textbf{K12Vista scores after SFT on Qwen3.5-2B-Base with full samples from each dataset.}}
\label{tab:k12vista}
\small
\setlength{\tabcolsep}{4pt}
\resizebox{\linewidth}{!}{
\begin{tabular}{l ccccc cccc}
\toprule
\multirow{2}{*}{\textbf{Model}} 
& \multirow{2}{*}{\textbf{Math}} 
& \multirow{2}{*}{\textbf{Physics}} 
& \multirow{2}{*}{\textbf{Chemistry}} 
& \multirow{2}{*}{\textbf{Biology}} 
& \multirow{2}{*}{\textbf{Geography}} 
& \multicolumn{4}{c}{\textbf{Overall}} \\
\cmidrule(lr){7-10}
& & & & & 
& \textbf{Choice} & \textbf{Blank} & \textbf{QA} & \textbf{Average} \\
\midrule
\multicolumn{10}{c}{\cellcolor{model_type}\textbf{Trained on Qwen3.5-2B-Base}} \\
\midrule
\textit{Qwen3.5-2B-Base}
& 81.18 & 79.24 & 80.41 & 79.48 & 77.64 & \cellcolor{column_green}83.65 & \cellcolor{column_green}76.19 & \cellcolor{column_green}79.09 & \cellcolor{column_green}79.72 \\
\textit{Qwen3.5-2B (Instruct)}
& 82.19 & 79.11 & 78.10 & 76.87 & 72.64 & 82.48 & 75.31 & 76.54 & 78.20 \\
\midrule
+ WizardLM (Full)
& 72.57 & 68.38 & 65.53 & 63.99 & 61.90 & 74.65 & 60.64 & 65.41 & 67.06 \\
+ DataFlow (Full)
& 71.80 & 77.12 & 70.70 & 76.17 & 67.75 & 77.64 & 66.77 & 67.21 & 72.87 \\
\midrule
+ K12-Train-Text
& 69.77 & 76.59 & 67.70 & 76.64 & 65.58 & 77.32 & 62.85 & 64.42 & 71.13 \\
+ K12-Train-MM
& 78.19 & 73.80 & 74.04 & 71.97 & 69.74 & 80.92 & 69.12 & 71.44 & 73.97 \\
\textbf{+ K12-Train-Full}
& 82.73 & 79.50 & 80.31 & 78.91 & 76.97 & \cellcolor{column_yellow}83.84 & \cellcolor{column_yellow}76.22 & \cellcolor{column_yellow}79.56 & \cellcolor{column_yellow}79.95 \\
\bottomrule
\end{tabular}}
\end{table*}

\paragraph{Gaokao-MM.}
Table~\ref{tab:gaokao-mm} evaluates multimodal educational reasoning across eight high-school subjects. K12-Train-Full achieves the highest overall accuracy (\textbf{39.9\%}) among all evaluated configurations, exceeding Qwen3.5-2B-Base and Qwen3.5-2B (Instruct) by 7.5\% and 3.8\%, respectively. Relative to the base model, K12-Train-Full improves performance in six of the eight subjects and matches it in another, with gains spanning both STEM and humanities disciplines. Its overall advantage is therefore broadly distributed rather than being driven by a single subject. Moreover, K12-Train-Full also surpasses K12-Train-Text and K12-Train-MM, providing initial evidence that textual curriculum supervision and multimodal grounding offer complementary benefits.

\paragraph{MDK12-Bench.}
Table~\ref{tab:mdk12} further evaluates multimodal educational reasoning across six science-oriented subjects. K12-Train-Full again achieves the highest overall score (\textbf{52.94}), surpassing Qwen3.5-2B-Base and Qwen3.5-2B (Instruct) by 2.17 and 2.33 points, respectively, and outperforming the strongest non-K12 baseline, DataFlow, by 1.66 points. Compared with the base model, it improves four of the six subject scores and achieves the best results among all configurations in Physics and Biology. It also exceeds K12-Train-Text and K12-Train-MM by 1.39 and 0.61 points, respectively, suggesting that the benefit of combining textual and multimodal supervision is not specific to Gaokao-MM.

\paragraph{K12Vista.}
Table~\ref{tab:k12vista} provides a particularly rigorous test of capability retention, as the untuned Qwen3.5-2B-Base already achieves a strong average score of 79.72. K12-Train-Full is the only fine-tuned configuration that improves upon the base model, reaching the best average score of \textbf{79.95}. Moreover, it achieves the best results across all three question formats---multiple-choice, fill-in-the-blank, and open-ended QA---showing that its advantage is not restricted to a particular response type. Its large gains over K12-Train-Text and K12-Train-MM further indicate that combining the two supervision sources is critical for preserving the base model's strong capabilities while improving its multimodal educational reasoning.

\subsubsection{Analysis}
\label{sec:exp_sft_analysis}
\paragraph{Why does K12-Train work?}
We hypothesize that the advantage of K12-Train stems from two properties of KG-guided synthesis. First, \emph{structural grounding}: each QA pair encodes not just a fact but an explicit \emph{relationship} (prerequisite, taxonomy, association, verification). Second, \emph{pedagogical coherence}: because the KG mirrors the official curriculum, the training distribution aligns with what educational benchmarks implicitly assume. Consistent with this, K12-Train's largest gains appear on open-ended GaokaoBench questions, EduEval's Ethics dimension (both of which demand structured reasoning rather than isolated recall), and K12Vista's open-ended QAs.

A further piece of evidence comes from \emph{cross-subject transfer}. Although K12-Train is synthesized exclusively from the mathematics, physics, chemistry, and biology subgraphs, SFT on K12-Train also attains the highest Chinese score (120.18, $+3.33$ over the strongest baseline DataFlow) and the best Humanities Math score (132.00, $+4.08$ over DataFlow), and matches the top tier on History, Geography, and Politics (Table~\ref{tab:gaokao-bench}). Since these subjects receive no in-domain supervision from K12-Train, the improvement cannot be attributed to content memorization and instead reflects the acquisition of a transferable, structurally-grounded answer style, supporting our claim that K12-Train teaches \emph{curriculum cognition} itself.

\paragraph{Complementarity of textual and multimodal supervision.}
K12-Train-Full achieves the best overall result on all three multimodal benchmarks and consistently outperforms both Text Only and MM Only, indicating that the two supervision sources are complementary. Text-only QA strengthens curriculum knowledge and relation-level reasoning, while multimodal QA grounds these capabilities in figures and diagrams. This pattern is consistent with prior findings that text-only instruction data can transfer reasoning and instruction-following capabilities to vision-language tasks~\citep{jia2025visualwebinstruct, lin2024vila, tu2025mlan}.

\section{Conclusion}
\label{sec:conclusion}

We introduced K12-KGraph, a curriculum-aligned knowledge graph built from official Chinese K--12 textbooks, together with K12-Bench (for evaluating curriculum cognition) and K12-Train (for KG-guided SFT). Experiments show that (1)~current LLMs lack robust curriculum cognition despite strong factual recall (46\% EM for a strong open-source model on K12-Bench); (2)~KG-guided synthesis is remarkably sample-efficient across both textual and multimodal settings: under a matched budget of approximately 2{,}300 samples, K12-Train-Text outperforms equally sized subsets of eight mainstream SFT corpora on GaokaoBench and EduEval, while K12-Train-Full outperforms substantially larger general-purpose datasets on three multimodal educational benchmarks; and (3)~textual and visual supervision are complementary: K12-Train-Full consistently outperforms both K12-Train-Text and K12-Train-MM across the three multimodal educational benchmarks.
% (3)~K12-Train-Full consistently surpasses both K12-Train-Text and K12-Train-MM on the multimodal benchmarks, demonstrating the complementarity of textual and visual supervision. 
Together, these results highlight the potential of curriculum-grounded data for building more capable educational language and vision-language models.

\clearpage

\bibliographystyle{plainnat}
\bibliography{main}

\clearpage

\beginappendix

\section{K12-KGraph Construction Details}
\label{app:kg}

\subsection{Node and Edge Attribute Specification}
\label{app:field_schema}
Table~\ref{tab:schema} defines the high-level ontology of K12-KGraph (nine node types and fourteen edge types), and Table~\ref{tab:node_field_schema} summarizes the detailed attribute specifications for each node type.

\begin{table*}[t]
\centering
\caption{\textbf{K12-KGraph schema:} nine node types and fourteen relation types with allowed source$\to$target connections.}
\label{tab:schema}
\small
\setlength{\tabcolsep}{3pt}
\resizebox{\linewidth}{!}{
\begin{tabular}{@{}lll@{}}
\toprule
\textbf{Category} & \textbf{Type} & \textbf{Description} \\
\midrule
\multirow{9}{*}{Nodes}
 & \texttt{Book} & Textbook entity (subject, grade, publisher) \\
 & \texttt{Chapter} & Chapter with title and ordering \\
 & \texttt{Section} & Section with title and ordering \\
 & \texttt{Concept(Cpt)} & Core knowledge concept with definition \\
 & \texttt{Skill(Skl)} & Transferable method or technique \\
 & \texttt{Experiment(Exp)} & Laboratory experiment with instruments \\
 & \texttt{Exercise(Exe)} & Practice problem with stem and answer \\
 & \texttt{Figure(Fig)} &  Single picture in a textbook \\
 & \texttt{VisualElement(Vis\_Ele)} &  Local visual elements in a figure that have educational value \\
\midrule
\multirow{14}{*}{Edges}
 & \texttt{is\_a} & Taxonomic subsumption (Concept$\to$Concept) \\
 & \texttt{prerequisites\_for(prereq)} & Learning dependency (Concept/Skill$\to$Concept/Skill/Experiment) \\
 & \texttt{relates\_to(rel\_to)} & Semantic association (Concept$\to$Concept) \\
 & \texttt{verifies(verif)} & Experimental validation (Experiment$\to$Concept) \\
 & \texttt{tests\_concept(tes\_cpt)} & Exercise assesses concept (Exercise$\to$Concept) \\
 & \texttt{tests\_skill(tes\_skl)} & Exercise assesses skill (Exercise$\to$Skill) \\
 & \texttt{contains\_visual\_element}
 & Figure composition (Figure$\to$VisualElement) \\
 & \texttt{illustrates}
 & Figure-level knowledge grounding (Figure$\to$Concept/Skill/Experiment) \\
 & \texttt{refers\_to}
 & Element-level knowledge grounding (VisualElement$\to$Concept/Skill/Experiment) \\
 & \texttt{requires\_figure}
 & Visual dependency (Exercise$\to$Figure) \\
 & \texttt{supports\_edge}
 & Visual evidence for textual relation (Figure$\to$Edge) \\
 & \texttt{appears\_in} & Knowledge located in section (Concept/Skill/Experiment/Figures$\to$Section) \\
 & \texttt{leads\_to(lea\_to)} & Chapter-level necessary learning order (Chapter$\to$Chapter) \\
 & \texttt{is\_part\_of} & Chapter hierarchy (Section$\to$Chapter, Chapter$\to$Book) \\
\bottomrule
\end{tabular}
}
\vspace{-2mm}
\end{table*}

\begin{table*}[t]
\centering
\caption{\textbf{Node-level attribute schema in K12-KGraph.}}
\label{tab:node_field_schema}
\resizebox{\linewidth}{!}{
\begin{tabular}{llll}
\toprule
\textbf{Node Type} & \textbf{Required Fields} & \textbf{Optional Fields} & \textbf{Constraints} \\
\midrule
    \texttt{Concept} &
    id, name, definition, importance &
    examples, aliases, formula, unit &
    importance $\in \{\text{understand}, \text{master}, \text{important}\}$ \\
    \texttt{Skill} &
    id, name, description &
    examples &
    must be generalizable (not exercise-specific) \\
    \texttt{Experiment} &
    id, name, instruments, is\_student &
    process, phenomena, conclusion &
    is\_student $\in \{0,1\}$ \\
    \texttt{Exercise} &
    id, stem, answer, difficulty, type &
    analysis &
    difficulty $\in [1,5]$, must link to Concept/Skill \\
    \texttt{Figure} &
    id, name, img\_path, description, textual\_evidence &
    role, role\_rationale &
    role is one of six pedagogical functions when annotated \\
    \texttt{VisualElement} &
    id, name, description, source\_figure &
    text\_on\_img, bbox\_2d, bbox\_conf, bbox\_rationale &
    bbox uses normalized $[y_{\min},x_{\min},y_{\max},x_{\max}]\in[0,1000]^4$ \\
\bottomrule
\end{tabular}
}
\end{table*}

\paragraph{Global Constraints.}
All textual node attributes must be explicitly supported by the source textbook content. Attributes should not introduce information beyond the provided text, and any extracted field must be verifiable from the corresponding source passage.
For visual nodes, \texttt{description} and localization attributes must be supported by visible image content, while \texttt{textual\_evidence} must quote the aligned textbook text verbatim. Every model-inferred visual relation includes a rationale and confidence score.

\subsection{Source Documents and Preprocessing Setup}
K12-KGraph is constructed from official textbook PDFs from the People's Education Press (PEP) curriculum, covering mathematics, physics, chemistry, and biology across primary, middle, and high school levels.
The source files are obtained from public repository\footnote{\url{https://github.com/TapXWorld/ChinaTextbook}}.

For document preprocessing, we directly adopt the official MinerU pipeline and recommended settings to convert PDFs into structured text. We use the provided CLI interface and default configuration without introducing any custom parsing rules or modifications.

\subsection{Prompt for KG Extraction}
\label{app:prompts}
A simplified version of the prompt template used for LLM-based knowledge graph extraction (Stage~3 of the pipeline) is demonstrated in Figure~\ref{fig:kg_prompt}. 
% The full prompt includes detailed schema definitions presented in Table~\ref{tab:schema} and Table~\ref{tab:node_field_schema}. In the implementation, this template is sent as a single user message without a separate system prompt.

In implementation, we construct K12-KGraph in two phases. We first extract the textual component section by section. The textual extraction template is sent as a single user message without a separate system prompt, and its full version additionally incorporates the detailed schema definitions in Table~\ref{tab:schema} and Table~\ref{tab:node_field_schema}. We then construct the multimodal component from the textbook images, their corresponding section text, and the previously extracted textual graph. This phase proceeds in three steps: (a) image-level analysis and knowledge-relevance filtering; (b) pedagogical-role classification and visual-element identification; and (c) visual-element localization and extraction of multimodal relations that link the image to the textual component.

\begin{figure*}[t]
\begin{tcolorbox}[
    colback=prompt_green!5!white,
    colframe=frame1,
    title=SINGLE-TURN USER PROMPT FOR TEXTUAL K12-KGRAPH EXTRACTION,
    fonttitle=\bfseries,
    colbacktitle=prompt_green!50!white,
    coltitle=frame1,
    boxrule=1.5pt,
    arc=5pt,
    boxsep=5pt,
    left=12pt,
    right=12pt,
    top=12pt,
    bottom=12pt
]
You are a knowledge graph expert proficient in K--12 textbook research and development, skilled at extracting structured knowledge from textbook texts. \\
Your task is to: read the provided \textbf{textbook section text}, extract nodes and edges according to the specified \textbf{Schema Definition}, and output the JSON result. \\

\textbf{Requirements:} 
\begin{itemize}[leftmargin=1.5em, itemsep=0.15em]
    \item This task \textbf{primarily focuses on building concepts/skills and the relationships between them}. Exercises are optional or only a few are required; quality over quantity is preferable. 
    \item Extract the \textbf{truly important and clearly presented} concepts/methods from this section of the textbook. Do not create content or include trivial details. 
    \item Skills should be \textbf{generalizable and applicable} within the subject, not specific problem-solving techniques used in a particular question. 
    \item For each edge, if there are supporting sentences in the original textbook text, please provide the original text. 
\end{itemize}

\vspace{0.5em}
\textbf{Output Format:}
\begin{itemize}[leftmargin=1.5em, itemsep=0.15em]
    \item Output only a single JSON object. Do not output any explanatory text, comments, or Markdown code block markers.
\end{itemize}

\vspace{0.5em}
\textbf{Input Text:}
\begin{itemize}[leftmargin=1.5em, itemsep=0.15em]
    \item \textbf{Textbook text:} \{Section\_Markdown\_Content\} 
\end{itemize}

\vspace{0.5em}
Now, please generate the knowledge graph JSON: \\
\end{tcolorbox}
\caption{\textbf{Simplified prompt template for KG (textual component) extraction.} The actual extraction prompt additionally includes the full graph schema definition and JSON output requirements.}
\label{fig:kg_prompt}
\vspace{-0.1in}
\end{figure*}

\begin{tcolorbox}[
    breakable,
    colback=prompt_green!5!white,
    colframe=frame1,
    title=PROMPT FOR MULTIMODAL K12-KGRAPH EXTRACTION (STEP 1),
    fonttitle=\bfseries,
    colbacktitle=prompt_green!50!white,
    coltitle=frame1,
    boxrule=1.5pt,
    arc=5pt,
    boxsep=5pt,
    left=10pt,
    right=10pt,
    top=10pt,
    bottom=10pt
]
\textbf{System prompt:} \\
You are a rigorous and conservative multimodal analysis assistant for Chinese textbooks. \\
You will receive a single image from a textbook and the original textbook text from the section containing that image. Your task is to combine the visible content of the image with the original textbook text, extract image-level information, and determine whether the image is worth including in the knowledge graph. \\

\textbf{Output fields.} 
\begin{enumerate}[leftmargin=1.5em, itemsep=0.15em]
    \item \texttt{name}: a short Chinese semantic name suitable for this image in the knowledge graph.
    \item \texttt{textual\_evidence}: the verbatim textbook text most directly related to the image.
    \item \texttt{description}: a Chinese description of the overall content directly visible in the image.
    \item \texttt{is\_kg\_relevant}: whether the image is worth including in the knowledge graph; the value must be \texttt{"yes"} or \texttt{"no"}.
    \item \texttt{irrelevant\_reason}: if the image is irrelevant, provide a brief reason in Chinese; otherwise, use an empty string.
\end{enumerate}

\vspace{0.5em}
\textbf{Rules for \texttt{textual\_evidence}.}
\begin{enumerate}[leftmargin=1.5em, itemsep=0.15em]
    \item The evidence must come directly from the original textbook text. Rewriting, summarization, synonym substitution, and independently adding content are strictly prohibited.
    \item Prefer the figure caption, excluding a standalone figure number; original body-text sentences directly corresponding to the image may also be included.
    \item One or more related excerpts may be copied and concatenated in their original order. Every excerpt must be verbatim textbook text; do not add words, rewrite content, or reorganize excerpts into a new sentence that does not occur in the textbook merely to improve coherence.
    \item If the textbook contains no text directly related to the image, use an empty string.
\end{enumerate}

\vspace{0.5em}
\textbf{Rules for \texttt{description}.}
\begin{enumerate}[leftmargin=1.5em, itemsep=0.15em]
    \item Describe only content directly visible in the image. Do not add extracurricular knowledge, information outside the textbook, or additional inference.
\end{enumerate}

\vspace{0.5em}
\textbf{Rules for \texttt{is\_kg\_relevant}.}
\begin{enumerate}[leftmargin=1.5em, itemsep=0.15em]
    \item Mark the image as \texttt{"no"} if it is merely decoration, a column icon, a layout element, a cover background, or has no explicit connection to the instructional content of the section.
    \item Mark the image as \texttt{"yes"} if it can be explicitly connected to a concept, skill, experiment, problem context, learning task, or textbook narrative in the section.
    \item Images that may be marked \texttt{"yes"} include, but are not limited to, conceptual illustrations, structural diagrams, experiment images, problem-solving aids, real-life context images, problem-introduction images, observation or discussion materials, positive examples, counterexamples, and incorrect demonstrations.
    \item Mark the image as \texttt{"no"} if it is only broadly related, its specific instructional function cannot be determined, and it cannot be stably linked to a knowledge object, problem context, or learning task.
\end{enumerate}

\vspace{0.5em}
\textbf{Output-format rules.}
\begin{enumerate}[leftmargin=1.5em, itemsep=0.15em]
    \item Output JSON only, with no additional explanation.
    \item The JSON object must contain the specified fields and no additional fields.
\end{enumerate}

\vspace{0.75em}
\textbf{User prompt:}

Please analyze this textbook image.
\begin{itemize}[leftmargin=1.5em, itemsep=0.15em]
    \item \textbf{Section information:} \{section\_metadata\}
    \item \textbf{Full original text of the section:} \{section\_text\}
    \item \textbf{Image-structure information:} \{figure\_hint\}
\end{itemize}
\end{tcolorbox}

\begin{tcolorbox}[
    breakable,
    colback=prompt_green!5!white,
    colframe=frame1,
    title=PROMPT FOR MULTIMODAL K12-KGRAPH EXTRACTION (STEP 2),
    fonttitle=\bfseries,
    colbacktitle=prompt_green!50!white,
    coltitle=frame1,
    boxrule=1.5pt,
    arc=5pt,
    boxsep=5pt,
    left=10pt,
    right=10pt,
    top=10pt,
    bottom=10pt
]
\textbf{System prompt:} \\
You are a rigorous assistant for classifying the instructional functions of Chinese textbook images and identifying visual elements. \\
You will receive a textbook image that has already been determined to be knowledge-relevant, its image-level analysis, and six valid \texttt{figure\_role} values. \\
Your tasks are:
\begin{enumerate}[leftmargin=1.5em, itemsep=0.15em]
    \item Determine the primary ``instructional function'' performed by the image in the textbook.
    \item Identify visual elements that can genuinely be localized within the image and have instructional significance.
\end{enumerate}

\vspace{0.5em}
\textbf{Rules for \texttt{figure\_role}.} \\
The valid \texttt{figure\_role} values are limited to the following six literal labels: explaining a concept;introducing a real-life context;demonstrating an experiment;assisting problem solving;summarizing and consolidating knowledge;presenting data. \\
Note that \texttt{figure\_role} is not a classification of the figure's format, image type, or content topic. \\
If the image is a ``structural schematic / knowledge-principle diagram / principle diagram / flowchart / schematic,'' \texttt{figure\_role} should usually be \texttt{explaining a concept}. If it is an ``experiment demonstration / experimental-apparatus diagram / experimental-process diagram,'' the role should usually be \texttt{demonstrating an experiment}. If it is a ``statistical chart / table / curve graph / bar chart,'' the role should usually be \texttt{presenting data}. If it is a ``problem diagram / geometric diagram / condition diagram,'' the role should usually be \texttt{assisting problem solving}. \\

\vspace{0.5em}
\textbf{Rules for \texttt{visual\_elements}.} \\
Retain only objects, regions, annotations, organs, devices, arrows, curves, text, and other elements that can genuinely be localized in the image and have instructional significance. Do not name abstract actions, generalized conclusions, or overall phenomena as visual elements. Each visual element should contain only \texttt{name}, \texttt{description}, and \texttt{text\_on\_image}; do not output \texttt{kind}. \\

\textbf{Output requirements.}
\begin{enumerate}[leftmargin=1.5em, itemsep=0.15em]
    \item Output JSON only.
    \item Do not output Markdown.
    \item \texttt{figure\_role} must copy one of the six valid values exactly.
    \item \texttt{figure\_role\_rationale} may explain the figure format, but must not use the figure format itself as the \texttt{figure\_role}.
    \item The JSON object must contain the specified fields and no additional fields.
\end{enumerate}

\vspace{0.75em}
\textbf{User prompt:} \\
Please analyze the primary instructional \texttt{figure\_role} of this textbook image and identify its visual elements. \\
\begin{itemize}[leftmargin=1.5em, itemsep=0.15em]
    \item \textbf{Section information:} \{section\_metadata\}
    \item \textbf{Full original text of the section:} \{section\_text\}
    \item \textbf{Image information:} \{figure\_info\}
\end{itemize}
\end{tcolorbox}

\begin{tcolorbox}[
    breakable,
    colback=prompt_green!5!white,
    colframe=frame1,
    title=PROMPT FOR MULTIMODAL K12-KGRAPH EXTRACTION (STEP 3),
    fonttitle=\bfseries,
    colbacktitle=prompt_green!50!white,
    coltitle=frame1,
    boxrule=1.5pt,
    arc=5pt,
    boxsep=5pt,
    left=10pt,
    right=10pt,
    top=10pt,
    bottom=10pt
]
\textbf{System prompt:} \\
You are a rigorous and conservative assistant for multimodal knowledge-graph extraction and visual-element localization from Chinese textbooks. \\
You will receive a textbook image already determined to be knowledge-relevant, the original textbook text from its section, the image-level analysis, the identified visual elements, and candidate nodes and edges from the existing textual knowledge graph for that section. \\
Your tasks are: \\
\begin{enumerate}[leftmargin=1.5em, itemsep=0.15em]
    \item Generate a bounding box only for each supplied visual element.
    \item Extract only incremental knowledge-graph relations that are directly supported by the image.
\end{enumerate}

\vspace{0.5em}
\textbf{General principles.}
\begin{enumerate}[leftmargin=1.5em, itemsep=0.15em]
    \item Links may use only the supplied candidate-node IDs. Do not invent new Concept, Skill, Experiment, or Exercise IDs.
    \item \texttt{supports\_edges} may select only from the supplied candidate \texttt{edge\_ref} values. Do not invent a new \texttt{edge\_ref}.
    \item If the image is relevant but cannot be stably linked to any candidate node or edge, return empty relations.
    \item When uncertain, output less rather than guessing.
\end{enumerate}

\vspace{0.5em}
\textbf{Evidence requirements.}
\begin{enumerate}[leftmargin=1.5em, itemsep=0.15em]
    \item Do not add relations using extracurricular common sense, background subject knowledge, completion from textbook context, or abstract inference.
    \item Only content directly visible in the image, or direct alignment between the image and \texttt{textual\_evidence}, counts as valid evidence.
    \item Do not output a relation if it relies primarily on ``this is usually true according to common sense'' rather than ``this is explicitly shown in the image.''
\end{enumerate}

\vspace{0.5em}
\textbf{Requirements for \texttt{illustrates}.} \\
Output \texttt{illustrates} only when the entire image is directly used to explain a candidate knowledge point. Do not output it when the connection is weak, indirect, or requires a long chain of reasoning.

\vspace{0.5em}
\textbf{Requirements for \texttt{visual\_elements}.} \\
Do not add, delete, or modify any supplied visual element. For each input \texttt{ve\_id}, return only its bounding-box coordinates, confidence, brief rationale, and \texttt{refers\_to} relations. Use normalized 0--1000 coordinates in the format $[y_{\min},x_{\min},y_{\max},x_{\max}]$. If a visual element cannot be stably localized, set its bounding box to \texttt{null} and confidence to 0.0. \\
Confidence must be a decimal between 0 and 1 and reflect the strength of the image evidence. A value of 1.0 means that the evidence in the image is highly direct and explicit and supports the localization with almost no interpretation; 0.5 means that some supporting evidence exists but a small amount of alignment or judgment is still required; and 0.0 means that the image provides no valid support. Do not add visual elements; return results one by one only for the supplied \texttt{ve\_id} values.

\vspace{0.5em}
\textbf{Requirements for \texttt{supports\_edges}.} \\
The threshold for \texttt{supports\_edges} is the highest. Output one only when the image can directly serve as visual evidence for a candidate edge. If the image supports only a node-level concept but does not directly support the source--target relation expressed by the candidate edge, do not output \texttt{supports\_edges}.

\vspace{0.5em}
\textbf{Requirements for \texttt{required\_by\_exercises}.} \\
The threshold for \texttt{required\_by\_exercises} is extremely high. Output one only when removing the image would make the exercise impossible to understand, make its meaning indeterminate, or prevent completion of a key reasoning step. If the image is merely helpful, assists understanding, or makes the exercise easier, but the exercise can still be completed without it, do not output the relation.

\vspace{0.5em}
\textbf{Output requirements.}
\begin{enumerate}[leftmargin=1.5em, itemsep=0.15em]
    \item Every \texttt{rationale} must explicitly state what visible content in the image supports the relation; do not provide vague explanations.
    \item Relation \texttt{confidence} must be a decimal between 0 and 1 and reflect the strength of the image evidence. A value of 1.0 means that the evidence is highly direct and explicit and supports the relation with almost no interpretation; 0.5 means that some supporting evidence exists but a small amount of alignment or judgment is still required; and 0.0 means that the image provides no valid support.
    \item Return an empty array for any field with no output content.
    \item Output JSON only, with no additional explanation.
    \item The JSON object must contain the specified fields and no additional fields.
\end{enumerate}

\vspace{0.75em}
\textbf{User prompt:}
Please localize the existing visual elements in this textbook image and extract incremental relations that can be linked to the existing knowledge graph. \\
\begin{itemize}[leftmargin=1.5em, itemsep=0.15em]
    \item \textbf{Section information:} \{section\_metadata\}
    \item \textbf{Full original text of the section:} \{section\_text\}
    \item \textbf{Image information:} \{figure\_info\}
    \item \textbf{Supplied visual elements:} \{visual\_elements\_json\} \\
    \item \textbf{Candidate nodes:} \{candidate\_nodes\} \\
    \item \textbf{Candidate edges:} \{candidate\_edges\} \\
    \item \textbf{Candidate exercises:} \{exercise\_catalog\} \\
\end{itemize}
\end{tcolorbox}
\captionof{figure}{
\textbf{Simplified three-step prompt sequence for constructing the
multimodal component of K12-KGraph.}
(a) Step~1 performs image-level analysis and knowledge-relevance
filtering.
(b) Step~2 identifies the pedagogical role of each retained figure and
extracts its instructional visual elements.
(c) Step~3 localizes these visual elements and extracts multimodal
relations linking the figure to the existing textual component.
}
\label{fig:mmkg_prompt}

\section{K12-Bench Construction Details and Evaluation Protocol}
\label{app:bench}

\subsection{Distractor Pool and Structural Sampling Rules}
\label{app:distractor}
\paragraph{Overview.}
We construct distractors through a structured, graph-driven pipeline rather than generating them from an LLM from scratch. For each benchmark instance, the generator first derives the question and gold answer set from a target relation or graph query, and then samples distractors from a multi-layer pool of structurally relevant but non-gold nodes. The overall design follows a ``near-to-far'' principle: candidates are first drawn from graph-local neighborhoods and only expanded to broader curriculum contexts when necessary, ensuring both plausibility and difficulty without introducing semantically valid alternatives.

\paragraph{Distractor construction pipeline.}
For each instance, distractors are constructed in three stages:
\begin{enumerate}
    \item \textbf{Candidate pooling:} multi-layer expansion from graph neighborhoods to broader curriculum-based pools;
    \item \textbf{Rule-based filtering:} removal of gold answers, surface-form duplicates, and task-invalid candidates;
    \item \textbf{Pedagogical filtering:} LLM-based filtering to discard weak, trivial, or controversial distractors.
\end{enumerate}
Candidates are finally deduplicated and stably ordered before option sampling.

\paragraph{Candidate pool construction.}
Candidate pools are expanded in layers. Early layers consist of structurally proximate nodes, while later layers draw from broader curriculum scopes such as the same section, chapter, book, subject-stage, or subject. Each layer is only used when previous layers are insufficient, and candidates already selected in earlier layers are removed.

Graph distance is defined over the undirected union of \texttt{relates\_to}, \texttt{is\_a}, and \texttt{prerequisites\_for}. Thus, references to 1-hop, 2-hop, or 3-hop neighborhoods correspond to distances in this combined structural graph rather than any single relation type. We also incorporate ``sibling'' structures induced by shared \texttt{is\_a} parents or shared \texttt{prerequisites\_for} targets, as such nodes are often pedagogically adjacent while remaining distinct from the correct answers.

In practice, this layered expansion yields an initial pool of approximately 8--20 raw candidates per instance prior to filtering, reducing the risk of brittle items with insufficient negative options.

\paragraph{Candidate ranking.}
Within each layer, candidates are ranked by semantic similarity using BAAI/bge-small-zh-v1.5~\cite{zhang2023retrieve}. For each candidate, the final score averages (i) similarity between the candidate text and the finalized question text, and (ii) the maximum similarity between the candidate text and any gold answer. Node representations use the \texttt{name} field, while \texttt{Exercise} nodes preferentially use their \texttt{stem}. Candidates are sorted within each layer by descending similarity before subsequent filtering.

\paragraph{Task-specific sampling rules.}
The exact construction of candidate pools varies by task family:
\begin{itemize}
    \item \textbf{\textsc{Ground} (exercise $\rightarrow$ concept/skill):}
    Distractors are seeded from the 2-hop neighborhoods of the correct answers and expanded to same-section, same-chapter, same-book, subject-stage, and subject-level pools with the same node type.
    \item \textbf{\textsc{Ground} (concept/skill $\rightarrow$ exercise):}
    The first layer consists of exercises attached to the 2-hop structural neighborhood of the query node, followed by broader curriculum-based expansions.
    \item \textbf{\textsc{Prereq}:}
    Candidate pools are asymmetric by design. For prerequisite queries, early layers emphasize forward descendants of the query node and the boundary of the gold prerequisite closure. For successor queries, early layers emphasize the full prerequisite closure and deeper descendants beyond the direct gold successors.
    \item \textbf{\textsc{Neighbor}:}
    The answer set consists only of direct \texttt{relates\_to} and \texttt{is\_a} neighbors. Distractors are drawn from the immediately outer ring, primarily 2-hop concept nodes and broader same-location concept pools.
    \item \textbf{\textsc{Evidence}:}
    First-layer distractors are constructed from structurally nearby concepts or experiments within the 2-hop neighborhood of the query or answer node, and then expanded through the same section/chapter/book/stage/subject hierarchy.
    \item \textbf{\textsc{Locate}:}
    For ``first appearance'' questions, distractors are drawn from alternative locations where the queried node appears and expanded to nearby sections or chapters. For chapter-level \texttt{leads\_to} tasks, the first layer contains nearby earlier chapters relative to both the query and gold prerequisite chapters, with fallback to broader chapter pools when necessary.
\end{itemize}

\paragraph{Final filtering.}
In addition to removing gold answers and name-level duplicates, task-specific rule filters exclude candidates that are too closely related to the gold set under the intended task semantics (e.g., nodes reachable via short alternative paths or still acceptable under a looser interpretation). The remaining candidates are then passed to the pedagogical filter described in the next subsection, ensuring that final distractors are both challenging and unambiguously incorrect.

\subsection{Prompt for Pedagogical Filtering}
To ensure that distractors are pedagogically valid and unambiguously incorrect, we apply an LLM-based filtering step after rule-based candidate pruning. The goal of this step is not to improve semantic similarity, but to remove candidates that are either trivially implausible or potentially acceptable as correct answers under a reasonable educational interpretation.

Each candidate is evaluated independently, given the question and the gold answer set. The model is instructed to act as a conservative K--12 teacher and determine whether the candidate is a valid distractor, as shown in Figure~\ref{fig:bench_filter_prompt}.

Candidates labeled as \texttt{INVALID} are removed from the pool. Only candidates labeled as \texttt{VALID} are retained for final option sampling.
\begin{figure*}[t]
\begin{tcolorbox}[
    colback=prompt_yellow!5!white,
    colframe=frame3,
    title=PROMPT FOR FILTERING K12-BENCH CANDIDATE DISTRACTORS,
    fonttitle=\bfseries,
    colbacktitle=prompt_yellow!45!white,
    coltitle=frame3!90!white,
    boxrule=1.5pt,
    arc=5pt,
    boxsep=5pt,
    left=12pt,
    right=12pt,
    top=12pt,
    bottom=12pt
]
\textbf{System Prompt:} \\
You are an experienced K--12 teacher. \\
Your task is to judge whether a candidate option is a \textbf{valid distractor} for a multiple-choice question. \\

\textbf{Criteria.} \\
A valid distractor must satisfy all of the following:
\begin{itemize}[leftmargin=1.5em]
    \item It is \textbf{incorrect} given the question and the gold answers.
    \item It is \textbf{plausible} and relevant (not obviously unrelated).
    \item It is \textbf{not partially correct}, not ambiguous, and would not be accepted as correct by a careful teacher.
\end{itemize}

\vspace{0.5em}
\textbf{Instruction.}
\begin{itemize}[leftmargin=1.5em, itemsep=0.15em]
    \item Be conservative: if a candidate could be considered correct under any reasonable interpretation, mark it as invalid.
\end{itemize}

\vspace{0.5em}
\textbf{Output format.} 
\begin{itemize}[leftmargin=1.5em, itemsep=0.15em]
    \item Output exactly one label: \texttt{VALID} or \texttt{INVALID}.
\end{itemize}

\vspace{0.75em}
\textbf{User Prompt:}
\begin{itemize}[leftmargin=1.5em, itemsep=0.15em]
    \item \textbf{Question:} \{question\}
    \item \textbf{Gold answers:} \{gold\_answers\}
    \item \textbf{Candidate option:} \{candidate\}
\end{itemize}
Is this a valid distractor? \\
\end{tcolorbox}
\caption{\textbf{Prompt for judging whether a candidate option can serve as a distractor.} The system prompt defines the teacher-style judgment criteria, and the user prompt supplies the question, the gold answer set, and the candidate option to be checked.}
\label{fig:bench_filter_prompt}
\vspace{-0.1in}
\end{figure*}

\subsection{Benchmark Composition Statistics}
\label{app:bench_composition}

Table~\ref{tab:bench_stats} reports the per-subtask sample counts and the graph relation probed by each subtask, complementing the high-level statistics quoted in the main text (\S\ref{sec:bench}).

\begin{table*}[t]
\centering
\caption{\textbf{K12-Bench statistics.} All tasks use multi-select format with four options.}
\label{tab:bench_stats}
\resizebox{\linewidth}{!}{
\begin{tabular}{@{}llrp{5.5cm}@{}}
\toprule
\textbf{Task} & \textbf{Subtask} & \textbf{\#Samples} & \textbf{Probed Relation} \\
\midrule
\textsc{Ground} (Knowledge Grounding) & Subtask 1 & 1{,}754 & Exercise $\to$ Concept/Skill \\
                                      & Subtask 2 & 2{,}096 & Concept/Skill $\to$ Exercise \\
\midrule
\textsc{Prereq} (Prerequisite Reasoning) & Subtask 1 & 2{,}565 & Prerequisite closure \\
                                         & Subtask 2 & 2{,}852 & Direct successors \\
\midrule
\textsc{Neighbor} (Neighbor Recommendation) & -- & 4{,}358 & \texttt{is\_a} + \texttt{relates\_to} \\
\midrule
\textsc{Evidence} (Experiment Evidence) & Subtask 1 & 769 & Concept $\to$ Experiment \\
                                        & Subtask 2 & 622 & Experiment $\to$ Concept \\
\midrule
\textsc{Locate} (Cross-Chapter Indexing) & Subtask 1 & 8{,}456 & First appearance location \\
                                         & Subtask 2 & 168 & Chapter prerequisites (\texttt{leads\_to}) \\
\midrule
\textbf{Total} & & \textbf{23{,}640} & \\
\bottomrule
\end{tabular}
}
\end{table*}

\begin{figure*}[t]
  \centering
  \includegraphics[width=1\linewidth]{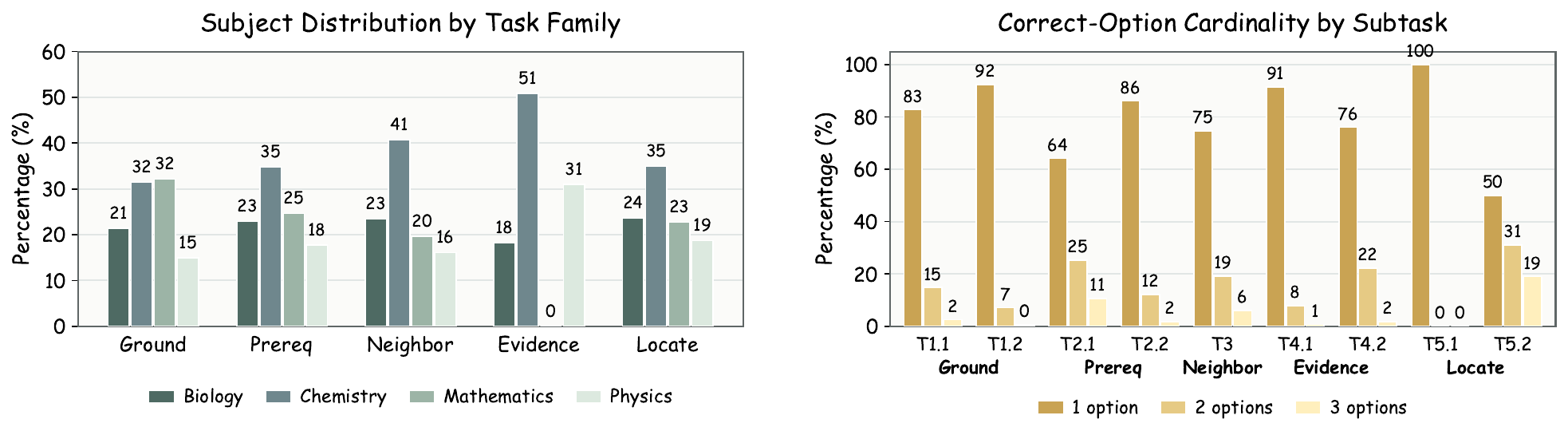}
  \caption{\textbf{Benchmark composition statistics for K12-Bench.} The left panel shows subject distribution across task families, and the right panel shows distribution of the number of correct options per subtask.}
  \label{fig:bench_distribution}
  \vspace{-0.1in}
\end{figure*}

Figure~\ref{fig:bench_distribution} summarizes two additional aspects of K12-Bench: (i) subject distribution across task families, and (ii) answer cardinality, i.e., the number of correct options per subtask. These properties are important for interpreting benchmark performance, as exact-match accuracy depends not only on reasoning difficulty but also on the structure of the answer space.

The left panel shows that subject coverage is broadly balanced across task families. For most tasks, items are evenly distributed across biology, chemistry, mathematics, and physics, without a clear bias towards any single subject. \textsc{Evidence} exhibits a different pattern: it draws primarily from chemistry, biology, and physics, and contains no mathematics items. This is a direct consequence of the graph schema, as mathematics does not include \texttt{Experiment} nodes and therefore cannot instantiate \texttt{verifies}-based queries. Overall, the subject distribution reflects the underlying curriculum structure rather than sampling artifacts.

The right panel shows that single-answer items dominate several task families, especially \textsc{Locate}. This pattern is largely induced by the semantics of the underlying graph queries. Many queries are structurally single-answer: for example, ``first appearance'' in \textsc{Locate} typically corresponds to a unique earliest location, and tightly constrained relation queries often yield a single valid node in the merged graph. More generally, a substantial portion of K12-Bench is derived from relations whose gold sets are inherently small, leading to a natural skew toward single-answer instances.

At the same time, the final option construction stage explicitly enforces balance over both answer cardinality and label combinations. The number of retained correct options per item is selected from the feasible set under a balancing constraint over $k \in \{1,2,3\}$, and the corresponding label combination (e.g., \texttt{A}, \texttt{A,B}, \texttt{B,C,D}) is assigned through a separate balancing procedure before populating slots \texttt{A}--\texttt{D}. As a result, while task families retain their intrinsic structural differences, the benchmark avoids systematic bias toward particular answer letters or specific label combinations.

\subsection{Answering Prompt and Decoding Rules}
All K12-Bench evaluations are run through the OpenAI-compatible chat interface. Each sample is formatted as a two-message conversation consisting of a fixed system prompt and a user prompt instantiated from the benchmark item. The combined prompt is shown in Figure~\ref{fig:bench_eval_prompt}. The system prompt constrains the model to answer using only option labels, while the user prompt presents the question stem and the four candidate options.

\begin{figure*}[t]
\begin{tcolorbox}[
    colback=prompt_purple!3!white,
    colframe=frame4,
    title=PROMPT FOR ANSWERING K12-BENCH QUESTIONS,
    fonttitle=\bfseries,
    colbacktitle=prompt_purple!40!white,
    coltitle=frame4!90!white,
    boxrule=1.5pt,
    arc=5pt,
    boxsep=5pt,
    left=12pt,
    right=12pt,
    top=12pt,
    bottom=12pt
]
\textbf{System Prompt:} \\
You are a K--12 teaching expert. \\

Please output the \textbf{set of correct option letters} according to the requirements of the question, and you must strictly adhere to the following format:
\begin{itemize}[leftmargin=1.5em, itemsep=0.15em]
    \item Only uppercase letters A/B/C/D are allowed;
    \item Output example: "A" or "A,C";
    \item Multiple letters must be arranged in ascending alphabetical order;
    \item Only English commas are allowed for separation; spaces are not permitted;
    \item \textbf{No explanations, analyses, prefixes, suffixes, punctuation, or line breaks are allowed}.
\end{itemize}

\vspace{0.5em}
Each question may have one or more correct options. \\
No points will be awarded for incorrect, multiple, or missing selections. \\

\vspace{0.75em}
\textbf{User Prompt:} \\
Question: \{question\} \\
A. \{A\} \\
B. \{B\} \\
C. \{C\} \\
D. \{D\} \\

Please provide the set of letters for the answer: \\
\end{tcolorbox}
\caption{\textbf{Prompt that instructs the model how to respond to K12-Bench questions.} The system prompt enforces an answer-only output format, and the user prompt presents the question together with the four labeled candidate options.}
\label{fig:bench_eval_prompt}
\vspace{-0.1in}
\end{figure*}

Across all released model configurations in \texttt{eval/configs/models/}, we
use deterministic decoding with \texttt{temperature} $=0.0$,
\texttt{top\_p} $=1.0$, and \texttt{max\_tokens} $=32$.

Model responses are post-processed by a lightweight parser to extract
predicted label sets. Both the raw model output and the parsed prediction are retained: the former is stored verbatim, while the latter is normalized (uppercased, deduplicated, and sorted) and used for evaluation.

The parser performs minimal normalization (e.g., unifying common delimiter variants) and extracts labels from the model output without attempting semantic repair. For abnormal cases (e.g., empty responses or API failures), instances are flagged and re-evaluated via automatic retries until a valid response is obtained.

\subsection{Baseline EM/F1 Computation for the Random Predictor}
\label{app:baseline_metrics}
For completeness, we formalize how exact match (EM) and option-label F1 are computed for the simple random baseline on K12-Bench. Let the gold label set for an instance be $G \subseteq \{A,B,C,D\}$ and the predicted label set be $\hat{G}$. Denote $k = |G|$ and $m = |\hat{G}|$. Per-instance scores are defined as
\begin{align}
\mathrm{EM}(G,\hat{G}) &= \mathbf{1}[G=\hat{G}], \\
\mathrm{P}(G,\hat{G}) &= \frac{|G \cap \hat{G}|}{|\hat{G}|}, \\
\mathrm{R}(G,\hat{G}) &= \frac{|G \cap \hat{G}|}{|G|}, \\
\mathrm{F1}(G,\hat{G}) &= \frac{2\,\mathrm{P}(G,\hat{G})\,\mathrm{R}(G,\hat{G})}{\mathrm{P}(G,\hat{G})+\mathrm{R}(G,\hat{G})},
\end{align}
with $\mathrm{F1}(G,\hat{G})=0$ when both precision and recall are zero.

\paragraph{Aggregation: instance-level (example-based) macro F1.}
To form a task- or benchmark-level F1, we average per-instance F1 across instances rather than pooling intersection/union counts:
\begin{equation}
\mathrm{F1}_{\mathrm{task}} \;=\; \frac{1}{N_{\mathrm{task}}}\sum_{i=1}^{N_{\mathrm{task}}} \mathrm{F1}(G_i,\hat{G}_i),
\end{equation}
and overall scores are obtained by weighting task-level averages by the task's instance count. This is the \emph{instance-level (example-based) macro F1}, equivalent to scikit-learn's \texttt{average=`samples'} convention for multi-label classification; every instance contributes equal weight regardless of its gold cardinality. We prefer this aggregation over a micro (corpus-pooled) F1 for two reasons: (i)~it keeps F1 on the same per-instance footing as EM, so EM and F1 numbers in the same row are directly comparable; and (ii)~micro-pooling would implicitly up-weight items with more correct labels, mixing difficulty with cardinality in an undesirable way for an evaluation aimed at curriculum cognition rather than label frequency.

For a random predictor, the baseline score is the \emph{expected} EM or F1 under the corresponding sampling rule. Equivalently, one may view the baseline as repeatedly sampling predictions and averaging the resulting scores over infinitely many trials.

\paragraph{Random guess.}
This baseline samples $\hat{G}$ uniformly from the $15$ non-empty subsets of $\{A,B,C,D\}$. For a fixed gold set $G$, the expected score is
\begin{align}
\mathbb{E}[\mathrm{EM}\mid G] &= \frac{1}{15}, \\
\mathbb{E}[\mathrm{F1}\mid G] &= \frac{1}{15}\sum_{\emptyset \neq S \subseteq \{A,B,C,D\}} \mathrm{F1}(G,S).
\end{align}
Because the label positions are symmetrically balanced, these expectations depend only on the gold cardinality $k=|G|$, not on the specific letters in $G$.

\subsection{Illustrative Example of KG-Grounded Resource Derivation}
Figure~\ref{fig:example} in the main text provides a concrete instantiation of the K12-KGraph-based construction pipeline. Both K12-Bench and K12-Train are generated by first selecting a target relation and extracting the corresponding local subgraph, followed by task-specific instantiation under structural constraints. The benchmark version preserves the original graph query form for evaluation, while the training version reformulates the same structure into instruction--response pairs for supervision.

\section{K12-Train Construction Details and SFT Protocol}
\label{app:train}

\subsection{Prompt for QA Synthesis}
\label{app:qa_prompts}
We use separate prompt templates for node-level and edge-level QA generation. As shown in Figure~\ref{fig:qa_prompt_node} and Figure~\ref{fig:qa_prompt_edge}, node-level prompts instruct the LLM to generate questions targeting the node's key attributes, while edge-level prompts instruct the LLM to generate questions that require reasoning about the relationship. For readability, Figure~\ref{fig:qa_prompt_edge} presents the shared logic of the edge-level prompt with a simplified input schema. In the implementation, each multimodal relation uses a relation-specific template supplied with the relevant figure or visual-element description, relation rationale, and, where applicable, the exercise stem or supported-edge metadata. The resulting VQA instance is paired with the complete figure for \texttt{illustrates}, \texttt{requires\_figure}, and \texttt{supports\_edge}, or with a copy of the complete figure in which the target bounding box is highlighted for \texttt{refers\_to}. Each synthesis template is sent as a single user message without a separate system prompt.

All prompts include style constraints: answers should be grade-appropriate in language, logically rigorous, and strictly grounded in the graph data. Before prompting, each node's property set is cropped to only the fields relevant for the target QA type (e.g., \texttt{importance} is removed for Concept QAs to prevent meta-commentary leakage).

\begin{tcolorbox}[
    colback=prompt_blue!5!white,
    colframe=frame2,
    title=SINGLE-TURN PROMPT FOR NODE-LEVEL QA SYNTHESIS,
    fonttitle=\bfseries,
    colbacktitle=prompt_blue!45!white,
    coltitle=frame2!80!white,
    boxrule=1.5pt,
    arc=5pt,
    boxsep=5pt,
    left=12pt,
    right=12pt,
    top=12pt,
    bottom=12pt
]
You are a top K--12 education expert with 20 years of experience, skilled at transforming complex STEM textbook knowledge into highly logical and thought-provoking QA materials. \\
Your task is to synthesize {n} high-quality SFT training datasets based on the provided Concept/Skill. 

\vspace{0.5em}
\textbf{Input Data}
\begin{itemize}[leftmargin=1.5em, itemsep=0.15em]
    \item \textbf{name:} \{name\}
    \item \textbf{properties:} \{properties\_json\}
\end{itemize}

\vspace{0.5em}
\textbf{Generation Strategy} \\
\textbf{1. General Constraints}
\begin{itemize}[leftmargin=1.5em, itemsep=0.15em]
    \item Each example must contain exactly \textbf{one question and one answer}.
    \item All content must be \textbf{strictly grounded in the input properties}; do not introduce external knowledge.
    \item Use aliases when appropriate to improve naturalness.
\end{itemize}

\textbf{2. Concept-Oriented QA}
\begin{itemize}[leftmargin=1.5em, itemsep=0.15em]
    \item Questions should focus on:
    \begin{itemize}[leftmargin=1.5em, itemsep=0.15em, label=\tiny$\blacksquare$]
        \item definition (what it is),
        \item key properties or formulas,
        \item basic understanding.
    \end{itemize}
    \item Answers must follow the structure:
    \[
    \backslash box\{\text{core definition}\}
    \;+\;
    \backslash box\{\text{core formula / key rule}\} \;(\text{optional})
    \]
    followed by:
    \begin{itemize}[leftmargin=1.5em, itemsep=0.15em, label=\tiny$\blacksquare$]
        \item 1~2 sentences of explanation (optional),
        \item 1 illustrative example (optional).
    \end{itemize}
\end{itemize}

\textbf{3. Skill-Oriented QA}
\begin{itemize}[leftmargin=1.5em, itemsep=0.15em]
    \item Questions should focus on:
    \begin{itemize}[leftmargin=1.5em, itemsep=0.15em, label=\tiny$\blacksquare$]
        \item what the method is,
        \item how to apply it (steps or procedure),
        \item when to use it (applicable scenarios).
    \end{itemize}
    \item Answers must follow the structure:
    \[
    \backslash box\{\text{method description or steps}\}
    \]
    followed by:
    \begin{itemize}[leftmargin=1.5em, itemsep=0.15em, label=\tiny$\blacksquare$]
        \item brief explanation of usage or advantages (optional),
        \item 1 illustrative example (optional).
    \end{itemize}
\end{itemize}

\vspace{0.5em}
\textbf{Style Requirements}
\begin{itemize}[leftmargin=1.5em, itemsep=0.15em]
    \item Language must be \textbf{clear, concise, and pedagogically appropriate}.
    \item Maintain a \textbf{formal and precise tone}, but avoid unnecessary complexity.
    \item Explanations should be accessible to K--12 students.
\end{itemize}
\end{tcolorbox}
\captionof{figure}{
\textbf{Node-level prompt for K12-Train QA synthesis.} Given a \texttt{Concept} or \texttt{Skill} and its properties, the model is asked to generate factually grounded question--answer pairs for K--12 learners.
}
\label{fig:qa_prompt_node}

\begin{tcolorbox}[
    breakable,
    colback=prompt_blue!5!white,
    colframe=frame2,
    title=SINGLE-TURN PROMPT FOR EDGE-LEVEL QA SYNTHESIS,
    fonttitle=\bfseries,
    colbacktitle=prompt_blue!45!white,
    coltitle=frame2!80!white,
    boxrule=1.5pt,
    arc=5pt,
    boxsep=5pt,
    left=12pt,
    right=12pt,
    top=12pt,
    bottom=12pt
]
You are a top K--12 education expert with 20 years of experience, skilled at transforming complex STEM textbook knowledge into highly logical and thought-provoking QA materials. \\
Your task is to synthesize \{n\} high-quality SFT training datasets based on the provided \textbf{edge relation} (\texttt{is\_a}, \texttt{prerequisites\_for}, \texttt{relates\_to}, \texttt{verifies}, \texttt{illustrates}, \texttt{refers\_to}, \texttt{requires\_figure}, \texttt{supports\_edge}). 

\vspace{0.5em}
\textbf{Input Data}
\begin{itemize}[leftmargin=1.5em, itemsep=0.15em]
    \item \textbf{source:} \{source\_name\}
    \item \textbf{target:} \{target\_name\}
    \item \textbf{properties:} \{properties\_json\}
\end{itemize}

\vspace{0.5em}
\textbf{Generation Strategy} \\
\textbf{1. General Constraints}
\begin{itemize}[leftmargin=1.5em, itemsep=0.15em]
    \item Each example must contain exactly \textbf{one question and one answer}.
    \item All content must be \textbf{strictly grounded in the input properties}; do not introduce external knowledge.
    \item Questions must focus on explaining \textbf{why the edge holds}, rather than merely restating the two endpoint names.
\end{itemize}

\textbf{2. Relation-Specific Prompting Goals}
\begin{itemize}[leftmargin=1.5em, itemsep=0.15em]
    \item \textbf{\texttt{is\_a}:}
    ask why \{source\_name\} belongs to or is part of \{target\_name\}; answers should explain how the source satisfies the defining properties of the target.
    \item \textbf{\texttt{prerequisites\_for}:}
    ask why one should learn \{source\_name\} before\{target\_name\}; answers should explain what knowledge or ability the source provides that supports later learning of the target.
    \item \textbf{\texttt{relates\_to}:}
    ask what relationship holds between \{source\_name\} and \{target\_name\}; answers should clarify their connection, contrast, or shared properties.
    \item \textbf{\texttt{verifies}:}
    ask what principle is verified by \{source\_name\}, or how \{source\_name\} verifies \{target\_name\}; answers should trace the path from experiment content and observations to the verified concept.
    \item \textbf{\texttt{illustrates}:}
    ask what concept, skill, or experiment is illustrated by the entire figure, why the figure illustrates \{target\_name\}, or how the figure helps explain \{target\_name\}; answers should explain how the overall visual content represents the definition, characteristics, structure, procedure, process, or phenomenon associated with the target.
    \item \textbf{\texttt{refers\_to}:}
    ask which concept, skill, or experiment a specific visual element corresponds to, why the element refers to \{target\_name\}, or what role it plays in understanding the target; answers should connect the localized element's visible features to the corresponding properties of the target.
    \item \textbf{\texttt{requires\_figure}:}
    ask why an exercise requires the figure, what information the figure provides for solving the exercise, or how the figure contributes to understanding the problem; answers should identify the conditions, data, structures, phenomena, or relations supplied by the figure and explain why the exercise cannot be fully understood or solved without them.
    \item \textbf{\texttt{supports\_edge}:}
    ask how the figure supports the relation between \{source\_name\} and \{target\_name\}, which visual information provides evidence for that relation, or why the figure can serve as visual evidence for the edge; answers should state the source--target relation and explain how the visible content directly supports it.
\end{itemize}

\textbf{3. Preferred Answer Structures}
\begin{itemize}[leftmargin=1.5em, itemsep=0.15em]
    \item \texttt{is\_a}: because \{source definition\}, \{target definition\}, and \{source satisfies target definition\}, therefore \{source belongs to target\}.
    \item \texttt{prerequisites\_for}: because learning the target requires \{specific knowledge/ability\}, and the source provides this knowledge/ability, therefore the source should be learned first.
    \item \texttt{relates\_to}: first state the concrete relationship between the two concepts, then briefly explain each side of the relation.
    \item \texttt{verifies}: from the experiment's setup, operation, or observed phenomenon, explain what result is seen and why that result supports the target concept.
    \item \texttt{illustrates}: because the figure presents \{main visual content\}, which reflects the \{definition/characteristics/structure/method/steps/process/phenomenon\} of \{target\_name\}, the figure provides a visual illustration of \{target\_name\}.
    \item \texttt{refers\_to}: because the visual element exhibits \{visible features\}, which correspond to the \{definition/characteristics/structure/method/steps/process/phenomenon\} of \{target\_name\}, the element refers to \{target\_name\}.
    \item \texttt{requires\_figure}: because solving or interpreting the exercise requires \{specific information to be identified/analyzed/calculated\}, and this information is provided by \{conditions/data/structures/phenomena/relations visible in the figure\}, the figure is necessary for understanding or answering the exercise.
    \item \texttt{supports\_edge}: first state the concrete \{edge\_type\} relation between \{source\_name\} and \{target\_name\}; then explain what visual information the figure presents and how that information directly supports the relation; finally conclude that the figure provides visual evidence for the edge.
\end{itemize}

\vspace{0.5em}
\textbf{Style Requirements}
\begin{itemize}[leftmargin=1.5em, itemsep=0.15em]
    \item Language must be \textbf{clear, concise, and pedagogically appropriate}.
    \item Maintain a \textbf{formal and precise tone}, but avoid unnecessary complexity.
    \item Explanations should be accessible to K--12 students.
\end{itemize}
\end{tcolorbox}
\captionof{figure}{
\textbf{Edge-level prompt for K12-Train QA synthesis.} Given a typed relation and its endpoint nodes, the model is asked to generate factually grounded question--answer pairs for K--12 learners.
}
\label{fig:qa_prompt_edge}

\subsection{Training Configuration}
\label{app:training_details}
Table~\ref{tab:sft_config} provides the full-parameter configuration for the text-only SFT experiments, while Table~\ref{tab:sft_config_mm} reports the LoRA configuration for the multimodal experiments. Within each experiment group, all models share an identical set of hyperparameters for fair comparison. Hyperparameters not explicitly listed follow the default settings of the training framework.

\begin{table*}[t]
\centering
\small
\setlength{\tabcolsep}{4pt}
\caption{\textbf{Full-parameter SFT configuration for the text-only experiments.}}
\begin{tabular}{ll}
\toprule
\textbf{Component} & \textbf{Setting} \\
\midrule
Backbones & Qwen3-4B-Base; Llama3.1-8B-Base \\
Framework & LLaMA-Factory (v0.9.5.dev0) \\
SFT Type & Full-parameter fine-tuning \\
Max Sequence Length & 32768 \\
Per-device Batch Size & 1 \\
Gradient Accumulation & 4 \\
Effective Batch Size & 32 (1 $\times$ 4 $\times$ 8 GPUs) \\
Optimizer & AdamW \\
Learning Rate & $5\times10^{-6}$ \\
Scheduler & Cosine decay \\
Warmup & 10\% steps \\
Epochs & 3 \\
Precision & bf16 \\
Distributed Training & DeepSpeed ZeRO-3 \\
Random Seed & 42 \\
Checkpoint Selection & Last checkpoint \\
Hardware & 8$\times$A100 (80GB) \\
Training Time & $\sim$0.4--1.7 hours per run \\
\bottomrule
\end{tabular}
\label{tab:sft_config}
\end{table*}

\begin{table*}[t]
\centering
\small
\setlength{\tabcolsep}{4pt}
\caption{\textbf{LoRA SFT configuration for the multimodal experiments.}}
\begin{tabular}{ll}
\toprule
\textbf{Component} & \textbf{Setting} \\
\midrule
Backbone & Qwen3.5-2B-Base \\
Framework & LLaMA-Factory (v0.9.5.dev0) \\
SFT Type & LoRA \\
Template & \texttt{qwen3\_vl\_nothink} \\
LoRA Configuration & rank 8, alpha 16, dropout 0.05 \\
LoRA Target Modules & All modules \\
Max Sequence Length & 4096 \\
Max Image Size & 262{,}144 pixels \\
Per-device Batch Size & 4 \\
Gradient Accumulation & 2 \\
Effective Batch Size & 64 (4 $\times$ 2 $\times$ 8 GPUs) \\
Learning Rate & $1\times10^{-4}$ \\
Scheduler & Cosine decay \\
Warmup & 10\% steps \\
Epochs & 3 \\
Precision & bf16 \\
Distributed Training & 8-GPU distributed data parallel \\
Data Workers & 16 preprocessing; 4 dataloader \\
Random Seed & 42 (framework default) \\
Checkpoint Selection & Last checkpoint \\
Hardware & 8$\times$A100 (80GB) \\
\bottomrule
\end{tabular}
\label{tab:sft_config_mm}
\end{table*}

\subsection{Baseline Subsampling and Fairness Controls}
\label{app:fairness}
For the text-only experiments, to ensure a fair comparison under a fixed data budget, we subsample 2{,}300 instances from each baseline dataset, matching the size of K12-Train.

Importantly, when metadata (e.g., source tags, task types, or length categories) is available, we perform stratified sampling to preserve the original distribution of these attributes. This ensures that the subsampled data remains representative of the full dataset rather than being biased toward a particular subset.

Our goal is to isolate the effect of \emph{data quality and structural grounding} under a controlled budget, rather than total data scale. Using a fixed-size and distribution-preserving sampling strategy allows us to attribute performance differences to dataset characteristics rather than volume or sampling artifacts.

We further verify that different random samples from the same dataset yield similar performance trends in pilot experiments, suggesting that our conclusions are not sensitive to a particular subset.

\section{Validation and Quality Assurance}
\label{app:validation}

\subsection{Validation Strategy, Annotator Background, and Ethics}
To ensure the reliability of K12-KGraph and its downstream resources, we adopt a tiered validation strategy centered on the knowledge graph (KG), which serves as the foundation for both K12-Bench and K12-Train.

The KG receives the most intensive validation, combining automatic structural checks with full human verification. In contrast, K12-Bench and K12-Train are validated via targeted manual spot-checks rather than full re-annotation. This difference is motivated by how these resources are constructed. K12-Bench instances (including both answers and distractors) are deterministically derived from graph structure rather than generated by LLMs, making their correctness largely reducible to the correctness of the underlying KG. K12-Train, as a supervision dataset, does not require perfect instance-level accuracy: minor imperfections are tolerable as long as the data remains factually grounded and pedagogically useful. Therefore, verifying consistency with the validated KG is sufficient for both resources.

All human validation is conducted by domain-qualified annotators. Since K12-KGraph covers four core disciplines (mathematics, physics, chemistry, and biology), we recruit subject specialists accordingly, with three annotators per
subject (12 in total). Annotators are experienced K--12 education practitioners, collectively covering primary, middle, and high school levels, ensuring reliable assessment of both fine-grained concepts and cross-stage curriculum relations.

Annotators are compensated at rates that meet or exceed the local minimum wage, in accordance with standard academic annotation practices. All data are derived from public educational materials and do not involve sensitive personal information.

\subsection{KG Validation: Structural Checks and Human Verification}

\paragraph{Automatic structural checks.}
We first perform automatic consistency checks on the \texttt{is\_a} and \texttt{prerequisites\_for} subgraphs, both expected to be directed acyclic graphs (DAGs). Cycles and structural inconsistencies are detected after graph construction and merging.

Detected conflicts are then reviewed by subject annotators, who determine whether to remove, modify, or retain edges based on curriculum correctness. In cases of ambiguity, decisions are made through discussion among annotators to
ensure consistency with standard curriculum progression.

\paragraph{Human verification.}
After resolving structural conflicts, we conduct full human validation over the entire graph. Each node and edge is independently annotated by three annotators within the same subject area. Disagreements are identified and resolved through joint review, leading to a consensus decision (retain, modify, or remove). The consensus label is taken as final.

\paragraph{Visual relation filtering.}
Each model-inferred visual relation includes a rationale and confidence score. Based on confidence-stratified inspection, we retain \texttt{refers\_to} and \texttt{illustrates} relations with confidence at least 0.85, and \texttt{requires\_figure} and \texttt{supports\_edge} relations with confidence at least 0.95. A predicted bounding box is retained only when it is valid and its localization confidence is at least 0.9. After relation filtering, we remove a \texttt{Figure} if it has no retained \texttt{illustrates}, \texttt{supports\_edge}, or incoming \texttt{requires\_figure} relation and contains no \texttt{VisualElement} with a retained \texttt{refers\_to} relation. Visual elements without a retained \texttt{refers\_to} relation and unused reified edge references are pruned together with their incident edges.

To quantify annotation reliability, we report inter-annotator agreement (IAA) before adjudication (Table~\ref{tab:iaa}). Agreement tends to be higher for structurally explicit relations (e.g., \texttt{is\_a}) and lower for semantically nuanced ones (e.g., \texttt{relates\_to}). After adjudication, the final graph achieves high precision.
\begin{table*}[t]
\centering
\small
\setlength{\tabcolsep}{5pt}
\caption{\textbf{Inter-annotator agreement (IAA) before adjudication, measured by Fleiss' $\kappa$, across subjects and node/edge types.}}
\label{tab:iaa}
\begin{tabular}{llccccc}
\toprule
\textbf{Group} & \textbf{Category} & \textbf{Math} & \textbf{Physics} & \textbf{Chemistry} & \textbf{Biology} & \textbf{Overall} \\
\midrule
\multirow{4}{*}{Node}
& \texttt{Concept}    & 0.84 & 0.86 & 0.85 & 0.83 & 0.85 \\
& \texttt{Skill}      & 0.81 & 0.83 & 0.82 & 0.80 & 0.82 \\
& \texttt{Experiment} & --   & 0.85 & 0.84 & 0.83 & 0.84 \\
& \texttt{Exercise}   & 0.78 & 0.80 & 0.79 & 0.77 & 0.79 \\
\midrule
\multirow{5}{*}{Edge}
& \texttt{is\_a}                & 0.90 & 0.92 & 0.91 & 0.89 & 0.91 \\
& \texttt{prerequisites\_for}   & 0.85 & 0.87 & 0.86 & 0.84 & 0.86 \\
& \texttt{relates\_to}          & 0.68 & 0.70 & 0.69 & 0.67 & 0.69 \\
& \texttt{verifies}             & --   & 0.82 & 0.83 & 0.81 & 0.82 \\
& \texttt{tests\_concept/skill} & 0.79 & 0.81 & 0.80 & 0.78 & 0.80 \\
\midrule
\textbf{Overall} & -- & 0.83 & 0.85 & 0.84 & 0.82 & 0.84 \\
\bottomrule
\end{tabular}
\end{table*}

\subsection{Spot-Check Validation of K12-Bench and K12-Train}
We validate K12-Bench and K12-Train via stratified manual sampling.

For K12-Bench, we perform stratified sampling across task families, subjects, and grade levels, and manually review $15\%$ instances in total. Each instance is checked for consistency between the question text, gold answer set, and distractor options with respect to the underlying graph structure. We find that $98.4\%$ of sampled instances are fully correct, while the remaining cases primarily involve minor issues such as phrasing ambiguity or borderline distractor quality, rather than semantic errors.

For K12-Train, we sample $10\%$ QA pairs from different construction categories (e.g., node-grounded and relation-grounded) and evaluate factual consistency, pedagogical appropriateness, and linguistic clarity. Among the sampled instances, $96.9\%$ are judged to be fully correct, with most errors attributable to surface-level issues rather than factual inconsistencies.

Across both resources, most identified issues are minor (e.g., phrasing or distractor similarity) rather than semantic errors, confirming that the validated KG provides a reliable basis for both benchmark construction and QA
synthesis.

\section{Extended Results and Sanity Checks}
\label{app:results}

\subsection{Full Benchmark Results}
Table~\ref{tab:bench_results_subtask} and Table~\ref{tab:bench_results_subject} provide fine-grained K12-Bench results omitted from the main text. Rather than repeating the task-family averages in Table~\ref{tab:bench_results}, we report per-subtask and per-subject breakdowns, highlighting the sources of variation behind the aggregate results.
\begin{table*}[t]
\centering
\caption{\textbf{Per-subtask K12-Bench results, reported in \%.} EM = exact match; F1 is computed at the option-label level.}
\label{tab:bench_results_subtask}
\small
\setlength{\tabcolsep}{3pt}
\resizebox{\linewidth}{!}{
\begin{tabular}{l cc cc cc cc cc cc cc cc cc}
\toprule
\multirow{2}{*}{\textbf{Model}}
& \multicolumn{2}{c}{\textsc{Ground}\_1}
& \multicolumn{2}{c}{\textsc{Ground}\_2}
& \multicolumn{2}{c}{\textsc{Prereq}\_1}
& \multicolumn{2}{c}{\textsc{Prereq}\_2}
& \multicolumn{2}{c}{\textsc{Neighbor}}
& \multicolumn{2}{c}{\textsc{Evidence}\_1}
& \multicolumn{2}{c}{\textsc{Evidence}\_2}
& \multicolumn{2}{c}{\textsc{Locate}\_1}
& \multicolumn{2}{c}{\textsc{Locate}\_2} \\
\cmidrule(lr){2-3} \cmidrule(lr){4-5} \cmidrule(lr){6-7} \cmidrule(lr){8-9}
\cmidrule(lr){10-11} \cmidrule(lr){12-13} \cmidrule(lr){14-15}
\cmidrule(lr){16-17} \cmidrule(lr){18-19}
& \textbf{EM} & \textbf{F1}
& \textbf{EM} & \textbf{F1}
& \textbf{EM} & \textbf{F1}
& \textbf{EM} & \textbf{F1}
& \textbf{EM} & \textbf{F1}
& \textbf{EM} & \textbf{F1}
& \textbf{EM} & \textbf{F1}
& \textbf{EM} & \textbf{F1}
& \textbf{EM} & \textbf{F1} \\
\midrule
\multicolumn{19}{c}{\cellcolor{model_type}\textbf{Open Source Models}} \\
\midrule
\textit{Meta-LLaMA-3-8B-Instruct} & 9.1 & 58.0 & 3.8 & 52.3 & 6.4 & 50.5 & 2.5 & 45.6 & 3.8 & 53.4 & 4.3 & 55.7 & 6.3 & 54.9 & 11.5 & 53.9 & 6.5 & 48.8 \\
\textit{GLM-4.7-Flash}           & 43.0 & 74.1 & 30.8 & 68.3 & 12.2 & 59.1 & 14.4 & 54.3 & 15.2 & 59.6 & 46.3 & 75.9 & 30.5 & 68.9 & 48.9 & 66.3 & 13.1 & 60.3 \\
\textit{Ministral-3-14B-Instruct}& 40.8 & 74.7 & 44.9 & 75.5 & 17.2 & 61.0 & 19.5 & 57.9 & 14.5 & 59.8 & 47.3 & 77.6 & 27.2 & 67.8 & 59.3 & 72.2 & 19.0 & 58.1 \\
\textit{Qwen3-32B}               & 44.2 & 76.8 & 48.9 & 77.6 & 17.0 & 62.3 & 16.9 & 58.6 & 14.6 & 60.3 & 49.5 & 78.6 & 32.0 & 70.8 & 72.1 & 75.9 & 18.5 & 59.0 \\
\textit{Gemma-4-31B-IT}          & 45.7 & 76.8 & 54.8 & 80.9 & 20.3 & 63.6 & 35.4 & 61.7 & 15.0 & 60.7 & 50.1 & 78.0 & 34.9 & 68.8 & 73.4 & 73.5 & 19.0 & 62.0 \\
\midrule
\multicolumn{19}{c}{\cellcolor{model_type}\textbf{Proprietary Models}} \\
\midrule
\textit{GPT-4o}                  & 28.9 & 71.1 & 31.5 & 70.1 & 9.9 & 60.1 & 8.7 & 55.1 & 10.1 & 57.9 & 37.5 & 74.3 & 26.2 & 68.8 & 56.4 & 72.6 & 12.5 & 58.0 \\
\textit{GPT-5-mini}              & 29.0 & 70.8 & 31.6 & 70.0 & 10.5 & 60.1 & 9.4 & 55.4 & 12.5 & 59.0 & 38.4 & 74.8 & 26.2 & 69.2 & 56.4 & 73.1 & 10.1 & 58.6 \\
\textit{GPT-5.2}                 & 43.9 & 76.4 & 56.6 & 81.5 & 13.5 & 60.9 & 21.9 & 58.3 & 13.1 & 60.0 & 46.9 & 77.7 & 34.9 & 69.1 & 71.2 & 71.7 & 17.3 & 61.7 \\
\textit{Gemini-2.5-Flash}        & 56.1 & 75.6 & 59.6 & 75.8 & 24.0 & 55.3 & 34.8 & 57.2 & 15.4 & 56.0 & 55.0 & 75.9 & 37.3 & 67.2 & 73.9 & 74.4 & 16.3 & 52.9 \\
\textit{Gemini-3-Flash}          & 42.5 & 75.5 & 80.9 & 89.9 & 26.6 & 56.1 & 42.1 & 60.1 & 33.4 & 63.5 & 55.9 & 76.7 & 36.8 & 67.1 & 82.7 & 82.9 & 33.9 & 67.6 \\
\bottomrule
\end{tabular}}
\end{table*}

\begin{table*}[t]
\centering
\caption{\textbf{Per-subject K12-Bench results, reported in \%.} EM = exact match; F1 is instance-level (example-based) macro F1. This table complements
Table~\ref{tab:bench_results_subtask} by showing how performance varies across
curriculum domains.}
\label{tab:bench_results_subject}
\setlength{\tabcolsep}{4pt}
\resizebox{0.88\linewidth}{!}{
\begin{tabular}{l cc cc cc cc cc}
\toprule
\multirow{2}{*}{\textbf{Model}}
& \multicolumn{2}{c}{\textbf{Biology}}
& \multicolumn{2}{c}{\textbf{Chemistry}}
& \multicolumn{2}{c}{\textbf{Mathematics}}
& \multicolumn{2}{c}{\textbf{Physics}}
& \multicolumn{2}{c}{\textbf{Overall}} \\
\cmidrule(lr){2-3} \cmidrule(lr){4-5} \cmidrule(lr){6-7} \cmidrule(lr){8-9} \cmidrule(lr){10-11}
& \textbf{EM} & \textbf{F1}
& \textbf{EM} & \textbf{F1}
& \textbf{EM} & \textbf{F1}
& \textbf{EM} & \textbf{F1}
& \textbf{EM} & \textbf{F1} \\
\midrule
\multicolumn{11}{c}{\cellcolor{model_type}\textbf{Open Source Models}} \\
\midrule
\textit{Meta-LLaMA-3-8B-Instruct} & 8.8 & 53.5 & 8.1 & 53.2 & 5.2 & 51.3 & 5.7 & 52.0 & 7.2 & 52.6 \\
\textit{GLM-4.7-Flash}            & 34.4 & 64.4 & 28.8 & 63.2 & 34.0 & 64.5 & 31.2 & 64.0 & 31.7 & 63.9 \\
\textit{Ministral-3-14B-Instruct} & 39.9 & 68.3 & 35.2 & 66.9 & 38.9 & 67.5 & 37.2 & 67.2 & 37.5 & \cellcolor{column_green}67.4 \\
\textit{Qwen3-32B}                & 45.4 & 70.8 & 39.8 & 68.7 & 44.2 & 69.8 & 42.5 & 69.1 & \cellcolor{column_green}42.6 & \cellcolor{column_yellow}69.5 \\
\textit{Gemma-4-31B-IT}           & 48.9 & 70.2 & 44.1 & 69.0 & 47.8 & 69.9 & 46.3 & 69.0 & \cellcolor{column_yellow}46.4 & \cellcolor{column_yellow}69.5 \\
\midrule
\multicolumn{11}{c}{\cellcolor{model_type}\textbf{Proprietary Models}} \\
\midrule
\textit{GPT-4o}                   & 33.2 & 67.2 & 29.2 & 65.2 & 31.9 & 65.9 & 31.3 & 65.6 & 31.1 & 65.9 \\
\textit{GPT-5-mini}               & 33.8 & 67.7 & 29.8 & 65.7 & 32.4 & 66.4 & 31.9 & 66.1 & 31.7 & 66.4 \\
\textit{GPT-5.2}                  & 45.5 & 69.2 & 39.9 & 67.2 & 44.8 & 68.5 & 42.7 & 67.4 & 42.8 & \cellcolor{column_green}68.0 \\
\textit{Gemini-2.5-Flash}         & 51.3 & 68.0 & 45.0 & 65.8 & 50.8 & 67.1 & 48.0 & 66.3 & \cellcolor{column_green}48.3 & 66.7 \\
\textit{Gemini-3-Flash}           & 60.8 & 74.5 & 53.6 & 72.0 & 59.8 & 73.7 & 56.2 & 72.3 & \cellcolor{column_yellow}57.1 & \cellcolor{column_yellow}73.0 \\
\bottomrule
\end{tabular}}
\end{table*}

\subsection{Stability Across Random Seeds}
\label{app:stability}
Although we do not conduct a full multi-seed evaluation on the entire benchmarks due to computational cost, we perform controlled multi-seed validation via stratified subsampling to assess robustness.

Specifically, for both GaokaoBench and EduEval, we randomly sample 20\% of instances \emph{within each subtask} to preserve the original task distribution. On this fixed subset, we repeat the SFT process with three different random seeds (\{42, 123, 2026\}) and evaluate the resulting models.

Table~\ref{tab:seed_stability} reports the overall performance on these sampled evaluation sets. We observe only minor variation across seeds, indicating that the performance gains are stable and not driven by a particular random initialization.

\begin{table*}[t]
\centering
\small
\setlength{\tabcolsep}{6pt}
\caption{\textbf{Performance variation across random seeds (\{42, 123, 2026\}).} Mean and standard deviation are computed over three runs on a 20\% stratified subset.}
\label{tab:seed_stability}
\begin{tabular}{lcc}
\toprule
\textbf{Backbone} & \textbf{GaokaoBench} & \textbf{EduEval} \\
\midrule
\textit{Qwen3-4B-Base}    
& 1002.94 $\pm$ 6.37 & 66.83 $\pm$ 0.11 \\
\textit{Llama3.1-8B-Base} 
& 621.18 $\pm$ 5.28  & 40.87 $\pm$ 0.23 \\
\bottomrule
\end{tabular}
\end{table*}

These results suggest that performance differences between datasets are stable with respect to seed choice.

\subsection{Overlap and Leakage Analysis}
We analyze potential data contamination between K12-Train and the external evaluation benchmarks (GaokaoBench and EduEval).

\paragraph{Source independence.}
K12-Train is synthesized from K--12 textbook content, focusing on curriculum structure and concept relations. In contrast, GaokaoBench consists of standardized high-stakes examination questions, while EduEval is constructed independently with diverse evaluation objectives. These sources are institutionally and functionally distinct.

\paragraph{Content characteristics.}
K12-Train emphasizes concept definitions, procedural knowledge, and relation explanations grounded in textbook structure, whereas GaokaoBench and EduEval primarily assess problem-solving and applied reasoning. The difference in content form and objective further reduces the likelihood of overlap.

\paragraph{Empirical verification.}
We perform n-gram overlap analysis between K12-Train and the evaluation benchmarks and observe negligible lexical overlap. Manual inspection of sampled instances also reveals no duplicated or near-duplicated question-answer pairs.

Taken together, these observations indicate that K12-Train does not introduce measurable leakage into the evaluation benchmarks.

\end{document}